\documentclass[lettersize,journal]{IEEEtran}
\usepackage{amsmath,amsfonts}
\usepackage{algorithmic}
\usepackage{algorithm}
\usepackage{array}
\usepackage[caption=false,font=normalsize,labelfont=sf,textfont=sf]{subfig}
\usepackage{textcomp}
\usepackage{url}
\usepackage{verbatim}
\usepackage{graphicx}
\usepackage{cite}
\usepackage{booktabs}
\usepackage{multirow}
\usepackage{threeparttable}
\usepackage{amssymb}
\usepackage[hidelinks]{hyperref}
\begin{document}

\title{Bi-EZP: LLM-Guided Bilevel Program Evolution for Ensemble Zero-Cost Proxy Discovery}

\author{Yutao Lai, Kezhao Lai, Hai-Lin Liu
        % <-this % stops a space

 \thanks{Yutao Lai, Kezhao Lai, and Hai-Lin Liu are with the School of Mathematics and Statistics, Guangdong University of Technology, Guangzhou 510006, China (e-mail: 2112114007@mail2.gdut.edu.cn; 3124006080@mail2.gdut.edu.cn; hlliu@gdut.edu.cn)
}}

\maketitle

\begin{abstract}
Zero-cost proxies enable neural architecture search (NAS) to rank candidate networks from statistics computed at initialization, avoiding the repeated training required by conventional performance estimation. Their efficiency, however, comes with a reliability limitation: different proxies measure different properties of an untrained network and often induce inconsistent rankings across search spaces. Ensemble proxies can combine complementary signals, but automated ensemble discovery must determine both a discrete aggregation structure and the continuous coefficients associated with that structure. Searching these variables jointly makes structural quality difficult to distinguish from parameter calibration. We propose Bi-EZP, a bilevel framework that separates the two decisions. At the upper level, a large language model generates executable aggregation programs over four complementary base proxies together with program-specific parameter bounds. At the lower level, covariance matrix adaptation evolution strategy (CMA-ES) optimizes the continuous parameters of each fixed program on an inner training split. The calibrated programs are compared using Kendall's rank correlation on a disjoint validation split, and evolutionary selection retains structures that generalize beyond their calibration data. The discovery phase is supervised by benchmark architecture accuracies, whereas the resulting frozen proxy evaluates new architectures without training them to convergence. Experiments on NATS-Bench and Network Design Spaces assess rank correlation across heterogeneous search spaces, while experiments in the DARTS space examine its use in downstream architecture search. The reported results show that explicitly separating program discovery from numerical calibration provides an effective route to automated ensemble zero-cost proxy construction within the evaluated settings. The source code is available at: \url{https://anonymous.4open.science/r/Bi-EZP-318D}.
\end{abstract}


\section{Introduction}

Neural architecture search (NAS) replaces hand-designed network construction with an optimization process over architectural choices~\cite{he2021automl}. Reinforcement-learning methods~\cite{zoph2016neural,pham2018efficient}, evolutionary search~\cite{real2017large,sun2020automatically}, and differentiable optimization~\cite{liu2018darts,xue2022partial,ye2022b} have all produced competitive architectures. Their progress has also exposed a persistent bottleneck: architecture evaluation dominates the search cost. Training every candidate is prohibitively expensive, while weight sharing reduces but does not remove the computational burden and can distort the relative quality of candidates.

Zero-shot NAS addresses this bottleneck by estimating architecture quality without full candidate training~\cite{li2024zero}. Zero-cost proxies extract statistics from a randomly initialized network using a small number of forward or backward passes. Parameter saliency~\cite{lee2018snip,wang2020picking}, activation patterns~\cite{mellor2021neural}, feature correlations~\cite{jiang2023meco}, and gradient consistency~\cite{li2023zico} provide inexpensive signals that can be reused by different NAS search algorithms. Yet these signals encode distinct inductive biases. Large-scale evaluations show that a proxy that is informative in one search space may be weak or even misleading in another~\cite{abdelfattah2021zero,ning2021evaluating}. The central difficulty is therefore not only to design another scalar heuristic, but to combine heterogeneous proxy evidence without committing to a brittle aggregation rule.

Existing ensemble and automated proxy methods provide two complementary routes. Fixed-form ensembles combine a prescribed set of signals through voting, ranking, or parameterized aggregation~\cite{lee2024az,huang2025evolving}. Symbolic approaches instead search over expressions or executable programs~\cite{akhauri2022eznas,wei2024auto,phan2025hand}. The former are numerically tractable but structurally restricted; the latter enlarge the structural space but introduce structure-dependent continuous parameters. This creates a coupled optimization problem. A useful program may appear weak when evaluated with poorly chosen coefficients, whereas extensive tuning can make a restricted structure appear competitive. Consequently, comparing uncalibrated structures does not isolate the value of the aggregation rule itself.

Recent advances in large language models (LLMs) have demonstrated capabilities beyond natural-language generation, particularly in optimization-related tasks that require structured reasoning and executable representations. LLMs have been investigated for generating optimization code~\cite{ma2026llamoco}, constructing black-box optimization benchmarks~\cite{wang2026evolution}, and directly producing candidate solutions for complex optimization problems~\cite{huang2025evaluation}. These studies suggest that LLMs can serve as flexible generators of structured optimization artifacts rather than merely text generators. This capability is particularly relevant to zero-shot proxy discovery, where the search object is not only a set of numerical coefficients but also a symbolic program specifying nonlinear transformations and interactions among multiple proxy signals. Compared with a manually predefined grammar, an LLM provides a more flexible structural prior that can generate diverse yet executable aggregation expressions.

Our key observation is that symbolic structure discovery and continuous parameter fitting play fundamentally different roles and should therefore be optimized at different levels. The symbolic structure determines which transformations and interactions among base proxies are expressible, whereas its parameter vector determines how a fixed structure behaves on a particular discovery sample. Bi-EZP formalizes this distinction as a bilevel optimization problem. The upper level searches executable aggregation programs together with their feasible parameter domains. For every proposed program, the lower level independently calibrates its numerical parameters before the program receives an upper-level fitness. This nested evaluation makes the comparison between candidate structures conditional on dedicated parameter optimization rather than on arbitrary initial coefficients.

Within this framework, Bi-EZP employs an LLM specifically as a structural generator rather than as a numerical optimizer or an autonomous NAS decision maker. Guided by prompts that encourage nonlinear transformations and cross-proxy interactions, the LLM proposes and varies candidate aggregation programs. A parser then constrains each response to a fixed four-proxy interface and an explicit set of finite parameter bounds, ensuring syntactic and numerical validity. Given a generated structure, CMA-ES optimizes its continuous parameters within these bounds, while validation Kendall's $\tau$ provides the upper-level fitness for tournament selection and elitist replacement. In this way, the LLM supplies flexible structural diversity, whereas CMA-ES and rank-based evolutionary selection provide explicit performance-driven optimization and validation.

The problem also differs from direct LLM-based architecture generation~\cite{zheng2023genius,chen2023evoprompting,nasir2024llmatic,cai2025seki}. Direct generation proposes one architecture at a time, whereas proxy discovery learns a reusable evaluation function that can score many candidates and can be embedded in different NAS procedures. The discovery phase uses benchmark accuracies as supervision; it is therefore not training-free in its entirety. Once the selected program and its parameters are frozen, however, evaluating another candidate requires only its initialization-time proxy statistics rather than full network training.

The main contributions of this work are summarized as follows:

\begin{itemize}

\item We formulate automated ensemble zero-cost proxy discovery as a 
bilevel optimization problem that explicitly decouples symbolic aggregation 
structure discovery from structure-dependent continuous parameter calibration. 
The upper level evolves executable proxy programs, while the lower level uses 
CMA-ES to optimize the continuous parameters of each fixed structure, enabling 
candidate structures to be compared after dedicated numerical calibration rather 
than under arbitrary coefficient settings.

\item We propose a parameterized prompting mechanism that enables LLMs to 
generate executable symbolic proxy structures together with their corresponding 
parameter constraints for lower-level numerical optimization. Each generated 
candidate jointly specifies an aggregation program and program-specific parameter 
bounds, which are validated through an executable-code checking and correction 
mechanism before being passed to the CMA-ES optimizer, thereby connecting open-ended LLM-based 
program generation with bounded numerical optimization.

\item We conduct extensive experiments on NATS-Bench, NDS, and DARTS across CIFAR-10, CIFAR-100, and ImageNet, where Bi-EZP demonstrates competitive performance in both rank-correlation evaluation and downstream architecture search.

\end{itemize}

\section{Related Work}

\subsection{Efficient Neural Architecture Search}

The computational cost of candidate evaluation has shaped the development of NAS. Early reinforcement-learning and evolutionary approaches train large numbers of sampled networks~\cite{zoph2016neural,real2017large,sun2020automatically}. Parameter sharing reduces this cost by evaluating subnetworks inside a common supernet~\cite{pham2018efficient}, while differentiable methods relax discrete architectural choices into continuous variables~\cite{liu2018darts,xue2022partial,ye2022b}. Other approaches use predictors, partial training, or population-based optimization to allocate evaluation effort more efficiently~\cite{cai2024eg}. These methods optimize architectures directly. Zero-shot NAS instead targets the evaluation function itself, replacing learned or trained performance estimates with statistics available near initialization.

\subsection{Zero-Cost Proxies}

Zero-cost proxies approximate architecture quality through inexpensive properties of an untrained network~\cite{abdelfattah2021zero,li2024zero}. Gradient-based criteria extend pruning-at-initialization signals such as SNIP and GraSP to architecture ranking~\cite{lee2018snip,wang2020picking}; GradNorm and related measures summarize the magnitude or organization of initialization gradients~\cite{abdelfattah2021zero}. Activation-based methods characterize the expressivity of randomly initialized mappings. NASWOT measures diversity in binary activation codes~\cite{mellor2021neural}, TE-NAS combines neural tangent kernel conditioning with linear-region counts~\cite{chen2021neural}, and Zen-NAS estimates expressivity through feature perturbations~\cite{lin2021zen}. More recent proxies exploit feature correlation, gradient consistency, or sample-wise activation patterns, including MeCo~\cite{jiang2023meco}, ZiCo~\cite{li2023zico}, and SWAP-NAS~\cite{peng2024swap}. L-SWAG further combines activation and gradient information for vision transformers~\cite{casarin2025swag}. Because these criteria observe different aspects of an initialized network, their rankings need not agree. Bi-EZP takes this heterogeneity as the input to ensemble discovery rather than assuming that any single signal is sufficient.

\subsection{Automated and Ensemble Proxy Discovery}

Ensemble proxies combine complementary signals to reduce dependence on a single heuristic. AZ-NAS integrates multiple zero-cost measures into a unified score~\cite{lee2024az}, while ParZC learns a parameterized representation for zero-cost prediction~\cite{dong2025parzc}. Per-architecture metric optimization provides another parameterized route to combining evaluation signals~\cite{linper}. These approaches show that calibration matters, but their parameterizations restrict the forms of interaction that can be expressed.

Automated symbolic discovery expands the candidate space. EZNAS evolves executable proxy expressions with genetic programming~\cite{akhauri2022eznas}, Auto-Prox searches training-free predictors for vision transformers~\cite{wei2024auto}, and symbolic regression has been used to derive performance-prediction formulas from primitive statistics~\cite{phan2025hand}. ECP is the closest setting to Bi-EZP because it combines NASWOT, MeCo, Sweet-SNIP, and SZiCo through a fixed power-function family and optimizes its coefficients with adaptive particle swarm optimization~\cite{huang2025evolving}. EvoREP~\cite{huang2025evorep} automatically evolves nonlinear ensembles of multiple zero-cost proxies to obtain more reliable architecture evaluations~\cite{huang2025evorep}. Bi-EZP retains this four-source basis but changes the optimization object: the aggregation program and its parameter bounds become outer-level variables, and each fixed program receives an independent inner numerical calibration. The distinction is therefore the separation of structural and parametric decisions, rather than a claim that one program-search formalism is intrinsically more compact than another.

\subsection{LLMs for NAS and Program Search}

Related work has explored several complementary ways of automating NAS with evolutionary optimization and LLMs. For example, LAPT~\cite{zhou2025design} transfers design principles between search tasks or incorporate rapid proxy feedback into LLM-guided architecture search~\cite{ji2025rz}. More recently, LLMENAS~\cite{lai2026llmenas} integrates LLM guidance into evolutionary NAS by dynamically adapting the search fitness according to historical optimization trajectories~\cite{lai2026llmenas}. These studies demonstrate the potential of both automated proxy composition and LLM-assisted evolutionary search. Bi-EZP connects these two directions but differs fundamentally in its search object: rather than evolving a predefined combination of proxies or using the LLM to guide the search for neural architectures, it employs the LLM to generate syntactically structured architecture-scoring programs. Explicit validation, CMA-ES-based continuous calibration, and rank-based evolutionary selection then determine whether a generated scoring program remains in the population.

\section{Methodology}

\subsection{Overview}

Bi-EZP searches for a reusable ensemble proxy over four precomputed zero-cost signals. The method contains two coupled but separately optimized levels. The upper level explores symbolic aggregation programs, using an LLM to initialize candidates and to produce crossover or mutation variants. The lower level receives one fixed program and optimizes only its continuous parameters with CMA-ES. After calibration on inner-training architectures, the candidate is evaluated on disjoint inner-validation architectures. Its validation Kendall's $\tau$ becomes the fitness used by the outer evolutionary loop.

\begin{figure*}[t]
  \centering
  \includegraphics[width=0.85\textwidth]{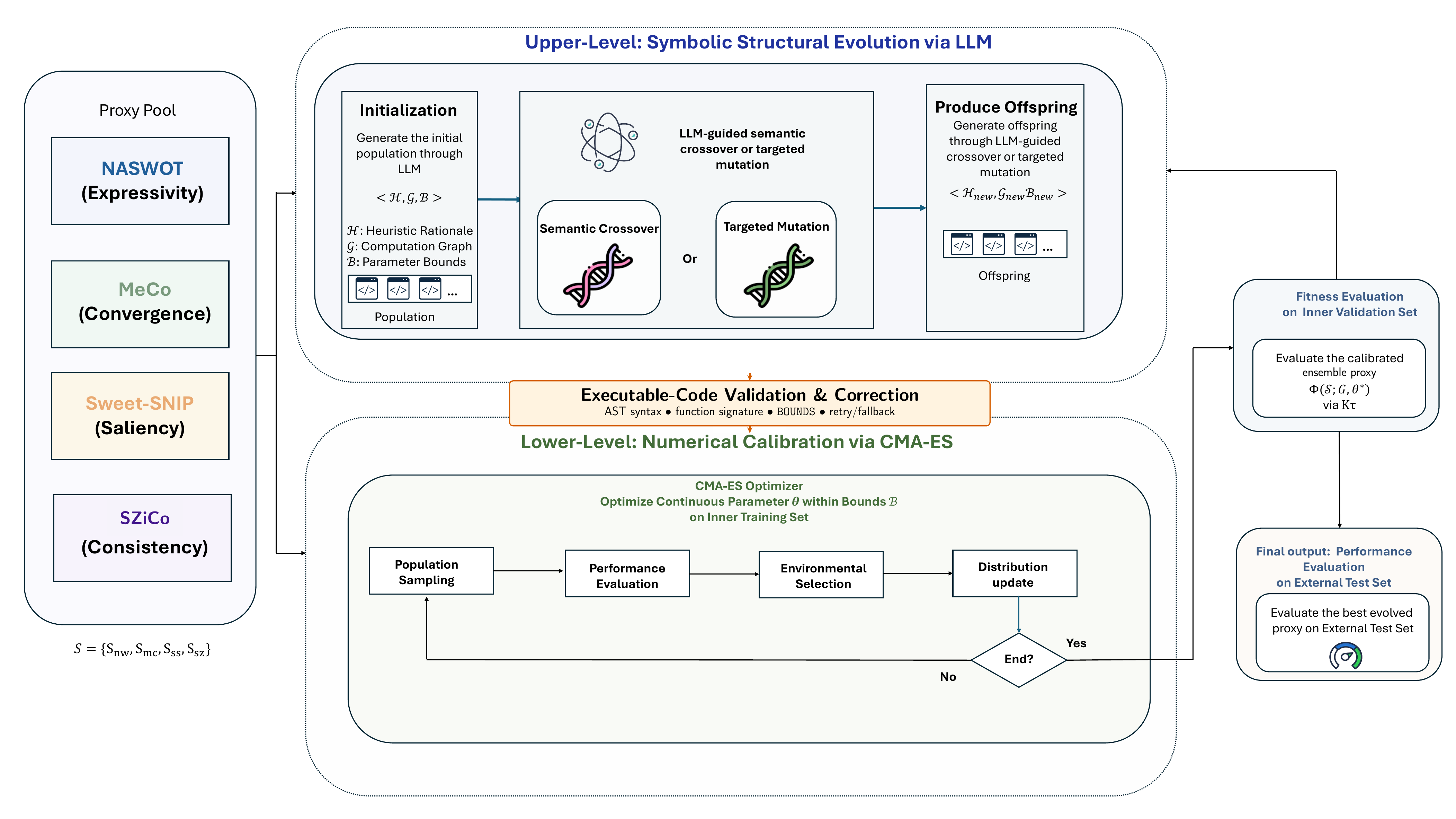}
  \caption{Overview of Bi-EZP. The upper level proposes executable aggregation structures and parameter domains. A code-validation gate checks syntax, interface, and bounds, applying correction attempts or a fixed fallback before a candidate reaches the lower-level CMA-ES calibration. Inner-training rank correlation drives numerical calibration, and inner-validation rank correlation drives structural selection.}
  \label{fig:framework}
\end{figure*}

Figure~\ref{fig:framework} illustrates this information flow. The LLM supplies a candidate description $\mathcal{H}$, executable structure $\mathcal{G}$, and finite parameter bounds $\mathcal{B}$. Before numerical calibration, an executable-code gate checks the abstract syntax tree, required function signature, and literal \texttt{BOUNDS} list. Invalid responses undergo a limited number of correction attempts; if all attempts fail, a fixed fallback program keeps the population evaluable. Only validator-accepted pairs $(\mathcal{G},\mathcal{B})$ define candidate-specific continuous domains $\Theta(\mathcal{G})$ for CMA-ES. The optimizer returns a calibrated vector $\theta^{*}$ inside that domain, after which the complete proxy $\Phi(\cdot;\mathcal{G},\theta^{*})$ is assigned a validation fitness. Thus, structural selection never compares unchecked or uncalibrated programs.

\subsection{Base Proxy Space}

The input basis is adopted from ECP~\cite{huang2025evolving} and is fixed as
\begin{equation}
    \mathcal{S}=\{s_{nw},s_{mc},s_{ss},s_{sz}\},
\end{equation}
where the four signals correspond to NASWOT, MeCo, Sweet-SNIP, and SZiCo, respectively. 
We select these four proxies to construct a compact yet heterogeneous input basis. 
They characterize an initialized architecture from complementary sources of network information: NASWOT captures activation-pattern diversity, MeCo describes feature correlation, Sweet-SNIP measures parameter saliency, and SZiCo reflects gradient consistency across input batches~\cite{huang2025evolving}. 
Together, they provide activation-, feature-, parameter-, and gradient-level views of network quality, reducing the dependence of the aggregation function on any single architectural characteristic. 
At the same time, these statistics can be obtained jointly using a forward pass and at most two backward propagations, which preserves the computational efficiency required by zero-shot NAS.

This paper keeps this proxy set fixed and focuses exclusively on discovering how these heterogeneous signals should be transformed and combined. 
The objective is therefore not to identify an optimal subset of zero-cost proxies, but to study whether automatically discovered transformations and interactions can yield a more effective architecture-scoring function than manually specified aggregation rules.

\subsubsection{NASWOT ($s_{nw}$)}
NASWOT provides the activation-level component of the basis and measures network expressivity through the diversity of binary activation patterns induced by a mini-batch~\cite{mellor2021neural}. For ReLU layer $r$, let $c_i^{(r)}$ denote the binary activation code of input $i$, $N_r$ the number of activation units, and $d_H$ the Hamming distance. The layer kernel and network score are
\begin{equation}
    K_{ij}^{(r)}=N_r-d_H\!\left(c_i^{(r)},c_j^{(r)}\right),
    \qquad
    s_{nw}=\log\left|\sum_{r=1}^{R}K^{(r)}\right|.
\end{equation}
The determinant summarizes how distinctly the initialized network partitions the sampled inputs. Unlike the remaining gradient- or feature-related signals, NASWOT depends only on activation patterns and therefore provides a complementary, gradient-free view of the architecture.

\subsubsection{MeCo ($s_{mc}$)}
MeCo provides a feature-level view by quantifying redundancy in intermediate representations through the minimum eigenvalue of a Pearson correlation matrix~\cite{jiang2023meco}. Following the formulation used by ECP~\cite{huang2025evolving}, four channels are sampled from the $C_l$ channels of layer $l$:
\begin{equation}
    s_{mc}=\sum_{l=1}^{L}\frac{C_l}{4}
    \lambda_{\min}\!\left(\mathbf{P}(\mathbf{f}_l)\right),
\end{equation}
where $\mathbf{f}_l$ contains the flattened sampled features and $\mathbf{P}(\cdot)$ denotes their Pearson correlation matrix. A larger minimum eigenvalue indicates that the sampled feature directions are less degenerate. Thus, MeCo complements activation-pattern information by describing the geometry and redundancy of learned feature representations, which have been associated with training convergence and generalization.

\subsubsection{Sweet-SNIP ($s_{ss}$)}
Sweet-SNIP supplies a parameter-saliency perspective. SNIP estimates the importance of an initialized parameter through the product between its magnitude and loss gradient~\cite{lee2018snip}. ECP further incorporates the sweet-gradient mechanism~\cite{yang2023sweet}, retaining gradients within a prescribed interval:
\begin{equation}
    s_{ss}=\sum_{l=1}^{L}
    \left|\boldsymbol{\theta}_l\circ
    \frac{\partial\mathcal{L}}{\partial\boldsymbol{\theta}_l}\right|
    \mathbb{I}\!\left\{
    \sigma_1\leq
    \left|\frac{\partial\mathcal{L}}{\partial\boldsymbol{\theta}_l}\right|
    \leq\sigma_2\right\}.
\end{equation}
Here, $\circ$ denotes element-wise multiplication and $\mathbb{I}\{\cdot\}$ is an indicator function. Whereas NASWOT and MeCo characterize network responses, Sweet-SNIP directly measures the sensitivity of the loss to initialized parameters, providing information about parameter importance that is not explicitly represented by the former proxies.

\subsubsection{SZiCo ($s_{sz}$)}
SZiCo provides a gradient-statistical perspective. ZiCo evaluates gradient consistency by comparing the mean gradient magnitude with its variation across input batches~\cite{li2023zico}. ECP combines this statistic with the sweet-gradient interval to obtain SZiCo~\cite{huang2025evolving}:
\begin{align}
    s_{zico}&=\sum_{l=1}^{L}\log\left(
    \sum_{\theta\in\boldsymbol{\theta}_l}
    \frac{\mathbb{E}_j[|\partial_{\theta}\mathcal{L}(\mathbf{X}_j,\mathbf{y}_j)|]}
    {\sqrt{\operatorname{Var}_j(\partial_{\theta}\mathcal{L}(\mathbf{X}_j,\mathbf{y}_j))}}
    \right),\\
    s_{sz}&=s_{zico}\,
    \mathbb{I}\!\left\{\sigma_1\leq
    \left|\frac{\partial\mathcal{L}}{\partial\theta}\right|
    \leq\sigma_2\right\}.
\end{align}
While Sweet-SNIP focuses on parameter-wise saliency, SZiCo measures the stability of gradient information across data batches and therefore captures a different aspect of trainability, convergence, and generalization.

\subsection{Parameterized Proxy Representation}
\label{sec:validity}

For architecture $i$, let
\begin{equation}
    \mathbf{s}_i=[s_{nw}^{(i)},s_{mc}^{(i)},s_{sz}^{(i)},s_{ss}^{(i)}]^{\top}
\end{equation}
denote its base-proxy vector. A candidate program maps a matrix of such vectors to one scalar score per architecture. We represent the candidate generated by the LLM as
\begin{equation}
    \mathcal{C}=\langle\mathcal{H},\mathcal{G},\mathcal{B}\rangle.
\end{equation}
The textual component $\mathcal{H}$ records the generated rationale, $\mathcal{G}$ is executable Python code implementing the aggregation rule, and
\begin{equation}
    \mathcal{B}=\prod_{k=1}^{d}[L_k,U_k],\qquad L_k<U_k,
\end{equation}
defines the feasible set $\Theta(\mathcal{G})$ of the $d$ continuous parameters used by that program. For a batch of $n$ architectures, the executable proxy is
\begin{equation}
    \Phi(\mathbf{S};\mathcal{G},\theta):
    \mathbb{R}^{4\times n}\times\Theta(\mathcal{G})
    \rightarrow \mathbb{R}^{n}.
\end{equation}

The executable interface turns open-ended generation into an evaluable search object. Every response must contain an ``Idea'' section and a ``Code'' section. The code must define a nonempty literal list \texttt{BOUNDS} and exactly one synchronous function
\begin{center}
\texttt{aggregate(s\_naswot, s\_meco, s\_zico, s\_ssnip, params)}.
\end{center}
The four proxy arguments are one-dimensional arrays with a shared architecture dimension; \texttt{params} supplies the values optimized by CMA-ES. The intended return value is one finite score vector of the same length.

Before evaluation, an abstract-syntax-tree parser checks Python syntax, the exact function signature, the absence of variable or keyword-only arguments, and a literal list of finite numerical bound pairs. Rejected responses are returned to the LLM together with the validation error for a limited number of correction attempts. If all attempts fail, the implementation inserts a fixed four-proxy weighted-sum candidate so that the population remains evaluable. Runtime failures, nonfinite outputs, and undefined rank correlations receive a penalty.

The resulting structural space is operational rather than a closed symbolic grammar: it is induced jointly by the prompt, the LLM, and the validator. The current parser constrains the interface and bounds but does not impose an operator whitelist, a maximum expression depth, or an explicit complexity penalty. This distinction matters because Bi-EZP searches executable programs under interface constraints; it does not enumerate a finite set of algebraic trees.

\subsection{Bilevel Discovery Objective}

Let the supervised discovery set be
\begin{equation}
    \mathcal{D}^{disc}=\{(\mathbf{s}_i,a_i)\}_{i=1}^{N},
\end{equation}
where $a_i$ is the benchmark accuracy of architecture $i$. The discovery pool is partitioned into disjoint inner-training and inner-validation subsets,
\begin{equation}
    \mathcal{D}_{train}=(\mathbf{S}_{train},\mathbf{A}_{train}),
    \qquad
    \mathcal{D}_{val}=(\mathbf{S}_{val},\mathbf{A}_{val}).
\end{equation}
The implementation uses an 80/20 split inside the sampled discovery pool. Architectures outside that pool form the subsequent evaluation set.

For fixed $(\mathcal{G},\mathcal{B})$, the lower level selects parameters that maximize rank agreement on $\mathcal{D}_{train}$:
\begin{equation}
    \theta^{*}(\mathcal{G},\mathcal{B})
    \in\arg\max_{\theta\in\mathcal{B}}
    \tau\!\left(
    \Phi(\mathbf{S}_{train};\mathcal{G},\theta),
    \mathbf{A}_{train}\right).
\end{equation}
The upper level evaluates the calibrated program on $\mathcal{D}_{val}$ and searches over the validator-accepted candidate space $\mathbb{Q}$:
\begin{equation}
\begin{split}
    (\mathcal{G}^{*},\mathcal{B}^{*})\in
    \arg\max_{(\mathcal{G},\mathcal{B})\in\mathbb{Q}}
    \tau\!\big(&\Phi(\mathbf{S}_{val};\mathcal{G},
    \theta^{*}(\mathcal{G},\mathcal{B})),\\
    &\mathbf{A}_{val}\big).
\end{split}
\end{equation}
Kendall's $\tau$ is used at both levels because the proxy is evaluated by the ordering it induces, not by calibrated prediction error. The two levels nevertheless receive different data and optimize different variables. The inner objective fits the numerical behavior of one fixed structure, whereas the outer objective compares calibrated structures by validation ranking. This separation prevents the outer loop from preferring a program merely because its initial coefficients happen to be favorable.

The bounds are part of the outer decision because different programs can require different parameter dimensionalities and numerical domains. Accordingly, a structural mutation may change the functional form, the number of parameters, their bounds, or several of these properties together. The inner solver is reinitialized for each accepted offspring; optimized parameters are not inherited across incompatible structures.

\subsection{LLM-Guided Structural Evolution}

At outer generation $t$, Bi-EZP maintains a population
\begin{equation}
    \mathbb{P}^{(t)}=
    \{\langle\mathcal{H}_i,\mathcal{G}_i,\mathcal{B}_i,
    \theta_i^{*},f_i\rangle\}_{i=1}^{P},
\end{equation}
where $f_i$ is the validation Kendall correlation of the calibrated program. Initialization requests $P$ independent candidates from a common prompt. Each valid candidate is calibrated on $\mathcal{D}_{train}$ before its first fitness is computed on $\mathcal{D}_{val}$, so the initial population follows the same nested evaluation protocol as later offspring.

Parent selection uses tournaments of size three among candidates whose fitness is valid. With probability $P_c$, two selected parents are inserted into a crossover prompt that requests a new program combining or modifying their mathematical components. Otherwise, one parent is inserted into a mutation prompt that requests a targeted structural change, such as a different nonlinear transformation or interaction. Both operators must return the same executable interface and a compatible \texttt{BOUNDS} list.

The numerical fitness does not appear directly in the crossover or mutation prompt. It influences generation through tournament selection: better validated structures are more likely to supply the code shown to the LLM. Likewise, the stored rationale $\mathcal{H}$ is retained for logging but is not supplied as persistent semantic memory to later prompts. The LLM therefore acts as a prompt-conditioned program proposal and variation operator. Structural survival is determined by external execution, CMA-ES calibration, and validation correlation.

Every generation produces $P$ offspring. After validation and lower-level calibration, parents and offspring are merged, sorted by $f_i$, and truncated to the best $P$ candidates. This elitist $(P+P)$ replacement retains the strongest programs found so far while maintaining continued exploration through newly generated code. The process terminates after a fixed number of outer cycles, and the highest-fitness calibrated candidate is returned.
\subsection{Lower-Level Parameter Adaptation via CMA-ES}
Once the LLM formulates the mathematical tuple $\langle \mathcal{H}, \mathcal{G}, \mathcal{B} \rangle$, the hypothesis space is strictly bounded. Identifying the optimal continuous coefficients within this specific parameter manifold $\Theta(\mathcal{G})$ is subsequently delegated to the Covariance Matrix Adaptation Evolution Strategy (CMA-ES)~\cite{hansen2003reducing}. We use CMA-ES because its derivative-free covariance adaptation is compatible with the rugged, non-differentiable, and potentially ill-conditioned rank-correlation objective.

The lower-level optimization is meticulously initialized based on the explicit priors $\mathcal{B}$ mapped by the LLM. For an $n$-dimensional parameter space bounded by the Cartesian product $\mathcal{B} = \prod_{i=1}^{n} [L_i, U_i]$, the initial search distribution mean $m^{(0)}$ is geometrically centered within the bounds, and the initial global step size $\sigma^{(0)}$ is scaled proportionally to the average boundary range:
\begin{equation}
    m^{(0)} = \frac{L + U}{2}, \quad \sigma^{(0)} = \max \left( \frac{1}{4n} \sum_{i=1}^{n} (U_i - L_i), 0.01 \right)
\end{equation}

At generation $g$, CMA-ES samples a population of $\lambda$ parameter vectors from a multivariate normal distribution parameterized by the current mean $m^{(g)}$ and covariance matrix $C^{(g)}$:
\begin{equation}
    \theta_{k}^{(g+1)} \sim m^{(g)} + \sigma^{(g)} \mathcal{N} \left( 0, C^{(g)} \right) \quad \text{for } k=1,\dots,\lambda
\end{equation}

To rigorously evaluate each sampled parameter vector $\theta_k$, the framework computes the composite proxy scores $\Phi(\mathcal{S}_{train}; \mathcal{G}, \theta_k)$ strictly on the inner-training set $\mathcal{D}_{train}$. Because LLM-generated symbolic expressions may occasionally produce undefined mathematical operations, we introduce a strict penalty mechanism. The objective function minimized by the lower-level solver is defined as:
\begin{equation}
    \mathcal{L}(\theta_k) = 
    \begin{cases} 
        - \tau \left( \Phi(\mathcal{S}_{train}; \mathcal{G}, \theta_k), \mathcal{A}_{train} \right), & \text{if } \Phi \text{ is valid} \\
        M, & \text{if } \Phi \text{ yields } \text{NaN/Inf}
    \end{cases}
\end{equation}
where $M$ is a large penalty scalar that immediately forces the evolutionary trajectory away from mathematically unstable parameter regions. Based on the $\mu$ best-performing parameter vectors, CMA-ES updates the covariance matrix $C^{(g+1)}$ using both rank-$\mu$ and rank-1 updates. The complete algorithmic implementation is formalized in Algorithm~\ref{alg:cmaes_lower}.

\subsection{Nested Search Procedure}

Algorithm~\ref{alg:ecp_llm} summarizes the interaction between the two optimization levels. The essential ordering is calibration before structural comparison. A newly generated program is first checked for syntactic and interface validity. CMA-ES then optimizes the parameters defined by its own bounds using only $\mathcal{D}_{train}$. The resulting parameter vector is frozen while the program receives its outer fitness on $\mathcal{D}_{val}$. Only this validation fitness participates in tournament selection and elitist replacement.

This ordering prevents validation information from entering the numerical objective of CMA-ES. It also ensures that a program is never carried into the population with an unevaluated parameter vector. If code generation fails repeatedly, the fallback candidate passes through the same lower- and upper-level evaluation path; it is not assigned a privileged fitness. Algorithm~\ref{alg:cmaes_lower} gives the lower-level procedure used for each accepted candidate.

\begin{algorithm}[htbp]
\caption{The Proposed Framework of Bi-EZP}
\label{alg:ecp_llm}
\textbf{Input:} $\mathcal{D}_{train}$, $\mathcal{D}_{val}$, population size $P$, outer cycles $T$, crossover probability $P_c$ \\
\textbf{Output:} $\langle\mathcal{H}^{*},\mathcal{G}^{*},\mathcal{B}^{*},\theta^{*}\rangle$
\begin{algorithmic}[1]
\STATE $\mathbb{P}^{(0)}\leftarrow\textsc{LLM-Initialize}(P)$
\FOR{each $\langle\mathcal{H}_i,\mathcal{G}_i,\mathcal{B}_i\rangle\in\mathbb{P}^{(0)}$}
    \STATE $(\mathcal{G}_i,\mathcal{B}_i)\leftarrow\textsc{Validate-Or-Correct}(\mathcal{G}_i,\mathcal{B}_i)$
    \STATE $\theta_i^{*}\leftarrow\textsc{CMA-ES}(\mathcal{G}_i,\mathcal{B}_i,\mathcal{D}_{train})$
    \STATE $f_i\leftarrow\tau(\Phi(\mathbf{S}_{val};\mathcal{G}_i,\theta_i^{*}),\mathbf{A}_{val})$
\ENDFOR
\FOR{$t=0$ to $T-1$}
    \STATE $\mathbb{O}\leftarrow\emptyset$
    \FOR{$j=1$ to $P$}
        \STATE $p_1\leftarrow\textsc{TournamentSelect}(\mathbb{P}^{(t)})$
        \IF{$\operatorname{rand}()<P_c$}
            \STATE $p_2\leftarrow\textsc{TournamentSelect}(\mathbb{P}^{(t)})$
            \STATE $c\leftarrow\textsc{LLM-Crossover}(\mathcal{G}_{p_1},\mathcal{G}_{p_2})$
        \ELSE
            \STATE $c\leftarrow\textsc{LLM-Mutation}(\mathcal{G}_{p_1})$
        \ENDIF
        \STATE $c\leftarrow\textsc{Validate-Or-Correct}(c)$
        \STATE $\theta_c^{*}\leftarrow\textsc{CMA-ES}(\mathcal{G}_c,\mathcal{B}_c,\mathcal{D}_{train})$
        \STATE $f_c\leftarrow\tau(\Phi(\mathbf{S}_{val};\mathcal{G}_c,\theta_c^{*}),\mathbf{A}_{val})$
        \STATE $\mathbb{O}\leftarrow\mathbb{O}\cup\{\langle c,\theta_c^{*},f_c\rangle\}$
    \ENDFOR
    \STATE $\mathbb{P}^{(t+1)}\leftarrow\textsc{TopP}(\mathbb{P}^{(t)}\cup\mathbb{O})$
\ENDFOR
\STATE \textbf{return} the highest-fitness member of $\mathbb{P}^{(T)}$
\end{algorithmic}
\end{algorithm}
\begin{algorithm}[htbp]
\caption{Lower-Level Parameter Calibration via CMA-ES}
\label{alg:cmaes_lower}
\textbf{Input:} Decoupled Sub-tuple $\langle \mathcal{G}, \mathcal{B} \rangle$ where $\mathcal{B} = \prod_{i=1}^{n} [L_i, U_i]$, Inner-training data $\mathcal{D}_{train} = \{\mathcal{S}_{train}, \mathcal{A}_{train}\}$, Max evaluations $E_{max}$ \\
\textbf{Output:} Optimal continuous parameters $\theta^*$
\begin{algorithmic}[1]
\STATE \textbf{Initialize:} 
\STATE Extract bounds $L, U$ from Cartesian product $\mathcal{B}$
\STATE $m \leftarrow (L + U) / 2$
\STATE $\sigma \leftarrow \max \left( \frac{1}{4n} \sum_{i} (U_i - L_i), 0.01 \right)$
\STATE $C \leftarrow \mathbf{I}_{n \times n}$ \hfill \textit{\% Initial covariance matrix}
\STATE $e \leftarrow 0$ \hfill \textit{\% Evaluation counter}
\WHILE{$e < E_{max}$ and not converged}
    \STATE Sample $\lambda$ offspring: $\theta_k \sim m + \sigma \mathcal{N}(0, C)$ for $k = 1, \dots, \lambda$
    \FOR{$k = 1$ to $\lambda$}
        \STATE $\theta_k \leftarrow \text{Clip}(\theta_k, L, U)$ \hfill \textit{\% Enforce explicit LLM boundaries $\mathcal{B}$}
        \STATE Compute proxy scores $\mathbf{s}_k = \Phi(\mathcal{S}_{train}; \mathcal{G}, \theta_k)$
        \IF{$\mathbf{s}_k$ contains NaN or Inf}
            \STATE $\mathcal{L}(\theta_k) \leftarrow 999.0$ \hfill \textit{\% Penalty for instability}
        \ELSE
            \STATE $\tau_k \leftarrow \text{KendallTau}(\mathbf{s}_k, \mathcal{A}_{train})$
            \STATE $\mathcal{L}(\theta_k) \leftarrow 1 - \tau_k$
        \ENDIF
        \STATE $e \leftarrow e + 1$
    \ENDFOR
    \STATE Sort the population such that $\mathcal{L}(\theta_{1:\lambda}) \le \mathcal{L}(\theta_{2:\lambda}) \dots \le \mathcal{L}(\theta_{\lambda:\lambda})$
    \STATE Update mean $m$ using the top $\mu$ vectors: $m \leftarrow \sum_{i=1}^{\mu} w_i \theta_{i:\lambda}$
    \STATE Update global step size $\sigma$ using cumulative step-size adaptation 
    \STATE Update covariance matrix $C$ using rank-$\mu$ and rank-1 updates
\ENDWHILE
\STATE \textbf{Return} $\theta^* \leftarrow \theta_{1:\lambda}$ \hfill \textit{\% Best evaluated parameters}
\end{algorithmic}
\end{algorithm}

\subsection{Discovery Protocol and Frozen-Proxy Evaluation}

Bi-EZP distinguishes proxy \emph{discovery} from subsequent proxy \emph{use}. Discovery is supervised: benchmark accuracies in $\mathbf{A}_{train}$ guide CMA-ES, and benchmark accuracies in $\mathbf{A}_{val}$ guide structural selection. Describing the complete discovery process as training-free would therefore be inaccurate. The zero-cost property applies after discovery, when $\mathcal{G}^{*}$ and $\theta^{*}$ have been fixed and the resulting proxy scores a candidate network from initialization-time statistics.

The evaluation protocol first samples a discovery pool from a benchmark search space, then partitions that pool into the inner-training and inner-validation subsets used by the two optimization levels. Architectures outside the discovery pool are reserved for final evaluation. For an architecture $x$ in this held-out set, evaluation computes the four base proxies and applies
\begin{equation}
    \hat{s}(x)=\Phi(\mathbf{s}(x);\mathcal{G}^{*},\theta^{*}).
\end{equation}
Neither the aggregation structure nor its parameters are updated during this stage. Cross-space or cross-dataset evaluation follows the same requirement: the complete proxy must be frozen before scores from the target setting are observed.

The distinction also clarifies the role of the DARTS experiments. Bi-EZP does not replace the architecture-search algorithm. It provides the search procedure with an evaluation signal that ranks candidate architectures without training each candidate to convergence. The cost of discovering this signal is incurred once, whereas the frozen aggregation program can subsequently score many architectures wherever the required base proxies are available.

\subsection{Computational Characteristics and Scope}

For population size $P$ and $T$ outer cycles, the procedure evaluates $P(T+1)$ candidate programs when every cycle produces a full offspring population. With at most $r$ generation or correction attempts per candidate, the number of LLM requests is bounded by $rP(T+1)$. If CMA-ES is limited to $E_{max}$ objective evaluations for each candidate, the inner level performs at most $E_{max}P(T+1)$ aggregation evaluations on the discovery training split. The total discovery cost therefore combines LLM inference, base-proxy extraction for the discovery architectures, repeated aggregation evaluation inside CMA-ES, and validation evaluation of calibrated programs.

The principal modeling assumption is that the four fixed base proxies contain complementary information that an aggregation program can exploit. Bi-EZP does not establish that this pool is optimal, nor does it penalize program complexity explicitly. Moreover, the executable search space depends on the language model, prompt, decoding behavior, and validator. Reproducible use therefore requires reporting the LLM version, decoding configuration, random seeds, population and cycle counts, correction and fallback rates, CMA-ES budget, data partitions, and the final generated program with its optimized parameters.

The current implementation validates syntax, interface shape, and finite parameter bounds, but it does not enforce a closed operator grammar or reject semantically duplicate programs. These choices favor structural flexibility at the cost of a less sharply characterized search space. The empirical claims in this work are consequently restricted to the supplied four-proxy basis, the reported benchmarks, and the evaluated discovery and transfer protocols.
\section{EXPERIMENTAL STUDIES}
\subsection{Datasets and Search Space}

To comprehensively evaluate the proposed framework and the discovered ensemble proxies, we conduct extensive experiments across diverse, well-established architectural search spaces and image classification datasets. 

\textbf{Search Spaces for Rank Correlation Evaluation.} To rigorously assess the effectiveness of the generated zero-cost proxies, we measure the Kendall's rank correlation coefficient ($\tau$) between the proxy scores and the ground-truth test accuracies. For this evaluation, we utilize two prominent benchmarks: NATS-Bench \cite{dong2021nats} and the Network Design Spaces (NDS) \cite{radosavovic2019network}. NATS-Bench provides a standardized environment containing both topology-based and size-based search spaces with fully trained ground-truth performances, allowing for precise, reproducible correlation analysis. NDS offers a broader set of network families (such as ResNet and ResNeXt variants), enabling us to test the structural generalization capacity of our composite proxies across highly heterogeneous architectural topologies.

\textbf{DARTS Search Space.} In addition to the benchmarks, we apply our automated proxy evaluation framework to the widely adopted DARTS search space. Evaluation within the continuous relaxation of DARTS tests whether the discovered proxy can serve as an architecture-ranking signal in a downstream search procedure.

\textbf{Datasets.} The evaluations across these search spaces are conducted on three standard visual recognition datasets: CIFAR-10, CIFAR-100 \cite{krizhevsky2009learning}, and ImageNet. CIFAR-10 and CIFAR-100 serve as the primary datasets for extracting initialized network statistics, conducting the structural evolution of the proxies, and performing the lower-level numerical calibration. To further verify the robust transferability and large-scale generalization of the architectures discovered via our optimized proxies, we extend our final evaluations to the high-resolution, large-scale ImageNet dataset.
\subsection{Implementation Details}
To ensure a fair and consistent comparison, the dataset configurations and evaluation protocols strictly follow the experimental settings established in prior ensemble proxy research~\cite{huang2025evolving}. Specifically, to construct the fitness evaluation datasets, we randomly sample 1,000 architectures along with their corresponding accuracies on the second training set for the NATS-Bench tasks. For the NDS search spaces, a subset of 500 architectures is sampled for fitness evaluation. To construct the fitness evaluation set within the open-domain DARTS search space, a random subset of 100 architectures and their corresponding accuracies on the CIFAR-10 second training set was sampled. Following the configuration in~\cite{huang2025evolving,xiang2023zero}, the search process on the DARTS space entails 10 search and 100 validation iterations. The discovered architectures are subsequently evaluated using the standard DARTS training pipeline~\cite{liu2018darts}. During the proxy extraction phase across all tasks, the input batch size is set to 128. Furthermore, for the gradient-based base proxies (i.e., SSNIP and SZiCo), the sweet gradient intervals are uniformly set to $[1e-4, 1e-3]$ for the NATS-Bench and NDS benchmarks. For the DARTS search space, these intervals are adjusted to $[1e-5, 1e-2]$ for CIFAR-10 and $[5e-5, 1e-4]$ for ImageNet.

For the proposed bilevel optimization framework, the upper-level structural evolution, which is driven by the Large Language Model (LLM), maintains a population size of 20 individuals. Specifically, GLM-4.7-Flash~\cite{glm2024chatglm} is employed as the underlying LLM in our experiments. The structural evolutionary search is conducted over a maximum of 20 generations. In the lower-level numerical calibration phase, the Covariance Matrix Adaptation Evolution Strategy (CMA-ES) is configured according to standard CMA-ES hyperparameter settings. The continuous parameter optimization strictly adheres to the boundaries generated by the LLM, with the maximum number of fitness evaluations for CMA-ES capped at 1,500 per structural hypothesis.

The implementation initializes the Python, NumPy, and PyTorch random-number generators with seed 1 before architecture sampling and proxy discovery. CMA-ES is invoked without a separate optimizer seed, and no deterministic seed is passed to the LLM API. LLM requests use the ZhipuAI v4 chat-completions endpoint with the model identifier \texttt{glm-4.7-flash}, temperature 1.0, disabled thinking, and non-streaming output. Top-p and the maximum output-token budget are not set explicitly and therefore follow the API defaults. Generated responses are parsed and checked using the validation and correction procedure described in Section~\ref{sec:validity}.
\begin{table}[htbp]
\centering
\caption{Score-Accuracy Correlation Comparison on NATS-Bench.}
\label{tab:correlation_comparison}
\begin{tabular}{lcccccc}
\toprule
\multirow{2}{*}{\textbf{Method}} & \multicolumn{3}{c}{\textbf{NATS-Bench-TSS}} & \multicolumn{3}{c}{\textbf{NATS-Bench-SSS}} \\
\cmidrule(lr){2-4} \cmidrule(lr){5-7}
& \textbf{C-10} & \textbf{C-100} & \textbf{IN-16} & \textbf{C-10} & \textbf{C-100} & \textbf{IN-16} \\
\midrule
ZiCo~\cite{li2023zico} & 0.604 & 0.593 & 0.589 & 0.709 & 0.536 & 0.719 \\
Zen-Score~\cite{lin2021zen} & 0.236 & 0.236 & 0.277 & 0.500 & 0.520 & 0.690 \\
SWAP~\cite{peng2024swap} & 0.641 & 0.664 & 0.613 & 0.502 & 0.233 & 0.430 \\
GradNorm~\cite{abdelfattah2021zero} & 0.466 & 0.474 & 0.429 & 0.524 & 0.338 & 0.512 \\
SynFlow~\cite{tanaka2020pruning} & 0.581 & 0.568 & 0.561 & 0.610 & 0.600 & 0.390 \\
ePADS~\cite{huang2025efficient} & 0.641 & 0.652 & 0.620 & 0.384 & 0.137 & 0.351 \\
NASWOT~\cite{mellor2021neural} & 0.604 & 0.622 & 0.601 & 0.430 & 0.184 & 0.403 \\
NI~\cite{wu2023training} & 0.508 & 0.517 & 0.481 & 0.722 & 0.468 & 0.667 \\
GradSign~\cite{zhang2021gradsign} & 0.619 & 0.600 & 0.593 & 0.210 & 0.160 & 0.040 \\
NTK~\cite{chen2021neural} & -0.349 & -0.364 & -0.321 & 0.200 & 0.618 & 0.423 \\
GraSP~\cite{wang2020picking} & 0.385 & 0.388 & 0.395 & -0.090 & 0.010 & 0.290 \\
SNIP~\cite{lee2018snip} & 0.472 & 0.474 & 0.433 & 0.420 & 0.460 & 0.570 \\
AZ-NAS~\cite{lee2024az}& 0.741 & 0.723 & 0.710 & 0.581 & 0.350 & 0.531 \\
\#Params & 0.576 & 0.552 & 0.519 & 0.190 & 0.210 & 0.380 \\
\#FLOPs & 0.541 & 0.517 & 0.487 & 0.530 & 0.540 & 0.650 \\
MeCo~\cite{jiang2023meco} & 0.730 & 0.711 & 0.669 & -0.601 & -0.717 & -0.581 \\
MeCoopt~\cite{jiang2023meco} & 0.691 & 0.712 & 0.689 & 0.674 & 0.641 & 0.617 \\
ParZC~\cite{dong2025parzc} & 0.706 & 0.743 & 0.699 & -- & -- & -- \\
$\xi$-GSNR~\cite{sun2023unleashing} & 0.661 & 0.658 & 0.608 & -- & -- & -- \\
EZNAS-A~\cite{akhauri2022eznas} & 0.650 & 0.650 & 0.610 & -- & -- & -- \\
ECP~\cite{huang2025evolving} & 0.782 & 0.771 & 0.740 & 0.771 & 0.722 & 0.806 \\
\midrule
\textbf{Bi-EZP} & \textbf{0.809} & \textbf{0.791} & \textbf{0.775} & \textbf{0.792} & \textbf{0.733} & \textbf{0.826} \\
\bottomrule
\end{tabular}
\end{table}

\begin{table}[htbp]
\centering
\caption{Score-Accuracy Correlation Comparison on NDS Search Spaces.}
\label{tab:nds_correlation}
\begin{tabular}{lccccc}
\toprule
\textbf{Method} & \textbf{DARTS} & \textbf{ENAS} & \textbf{PNAS} & \textbf{NASNet} & \textbf{Amoeba} \\
\midrule
ZiCo~\cite{li2023zico} & 0.345 & 0.197 & 0.195 & 0.087 & -0.019 \\
SWAP~\cite{peng2024swap} & 0.446 & 0.345 & 0.342 & 0.266 & 0.138 \\
GradSign~\cite{zhang2021gradsign} & 0.537 & 0.424 & 0.396 & 0.290 & 0.250 \\
NASWOT~\cite{mellor2021neural} & 0.480 & 0.387 & 0.363 & 0.299 & 0.208 \\
ePADS~\cite{huang2025efficient} & 0.507 & 0.485 & 0.429 & 0.404 & 0.374 \\
SynFlow~\cite{tanaka2020pruning} & -0.001 & -0.092 & -0.090 & -0.191 & -0.001 \\
EZNAS-A~\cite{akhauri2022eznas} & 0.560 & 0.520 & 0.510 & 0.440 & 0.450 \\
\#FLOPs & 0.500 & 0.413 & 0.395 & 0.288 & 0.238 \\
\#Params & 0.493 & 0.411 & 0.387 & 0.289 & 0.241 \\
AZ-NAS~\cite{lee2024az} & 0.416 & 0.446 & 0.375 & 0.396 & 0.368 \\
NI~\cite{wu2023training} & 0.305 & 0.327 & 0.268 & 0.278 & 0.114 \\
$\xi$-GSNR~\cite{sun2023unleashing} & 0.547 & 0.440 & 0.403 & 0.313 & 0.226 \\
ParZC~\cite{dong2025parzc} & 0.503 & 0.506 & -- & 0.385 & -- \\
MeCo~\cite{jiang2023meco} & 0.204 & 0.122 & 0.095 & 0.072 & 0.102 \\
MeCoopt~\cite{jiang2023meco} & 0.486 & 0.403 & 0.375 & 0.314 & 0.188 \\
GradNorm~\cite{abdelfattah2021zero} & 0.227 & 0.055 & 0.109 & 0.080 & -0.116 \\
ECP~\cite{huang2025evolving} & 0.568 & 0.512 & 0.476 & 0.468 & 0.406 \\
\midrule
\textbf{Bi-EZP} & \textbf{0.621} & \textbf{0.582} & \textbf{0.551} & \textbf{0.519} & \textbf{0.511} \\
\bottomrule
\end{tabular}
\end{table}
\subsection{Experimental Analysis}
\label{sec:experimental_analysis}

This section evaluates Bi-EZP across established neural architecture search benchmarks and open-domain search spaces. The experiments examine rank correlation, transfer across datasets and search spaces, sensitivity to design choices, and use of the proxy in downstream architecture search.

\subsubsection{Experimental Result in Rank Correlation Across NAS Benchmarks}
The Kendall’s rank correlation coefficient ($\tau$) serves as the primary metric for evaluating the ranking fidelity of zero-cost proxies, measuring the consistency between proxy predictions and ground-truth test accuracies. Table~\ref{tab:correlation_comparison} presents a comprehensive comparison on the NATS-Bench benchmark, covering both topology-based (TSS) and size-based (SSS) search spaces.

Bi-EZP achieves the highest rank correlation among the compared methods under the reported protocol. Relative to ECP, Bi-EZP increases the TSS $\tau$ score from $0.782$ to $0.809$ on CIFAR-10, from $0.771$ to $0.791$ on CIFAR-100, and from $0.740$ to $0.775$ on ImageNet-16-120. On SSS, the corresponding correlations increase from $0.771$ to $0.792$ on CIFAR-10 and from $0.806$ to $0.826$ on ImageNet-16-120.

ECP already employs an automated symbolic-regression mechanism for proxy discovery. These results are consistent with a benefit from separating symbolic structure search and continuous parameter calibration under the reported protocol. The bilevel formulation is intended to reduce the coupling between structural comparison and coefficient calibration; the comparison does not establish that the gain arises from program representation, operator choice, or expression size.

Bi-EZP also obtains higher rank correlations than the reported AZ-NAS and EZNAS-A results. These comparisons establish an empirical difference under the reported protocol, but they do not isolate program representation, operator choice, or expression size as its cause.
\subsubsection{Experimental Result in Rank Correlation Across NDS Benchmarks}
To further examine the structural generalization capability of the discovered proxies, we extend the evaluation to the Network Design Spaces (NDS) benchmark, which encompasses highly heterogeneous architectural families, including DARTS, ENAS, PNAS, NASNet, and Amoeba. Compared with NATS-Bench, the NDS benchmark presents significantly greater variability in architectural topology, thereby constituting a more stringent test of cross-space robustness.

As summarized in Table~\ref{tab:nds_correlation}, Bi-EZP obtains the highest rank correlation among the compared methods across the five evaluated architectural families. Relative to ECP, the correlation increases from $0.568$ to $0.621$ on DARTS, from $0.512$ to $0.582$ on ENAS, and from $0.406$ to $0.511$ on Amoeba. These per-space results establish an empirical improvement under task-specific discovery, while the frozen source--target evaluation below examines whether a discovered proxy can be reused without recalibration.

The per-space comparisons alone do not rule out search-space-specific fitting. Cross-space reuse is therefore assessed separately by freezing both the aggregation program and its calibrated parameters before transfer.

\subsubsection{Cross-Dataset and Cross-Search-Space Proxy Performance}

The preceding comparisons evaluate proxies discovered separately for each target. We further consider a stricter transfer setting, where the complete proxy discovered on a source task, including both its aggregation program and CMA-ES-optimized parameters, is frozen and directly evaluated on other targets. Rows in Tables~\ref{tab:tss_transfer} and~\ref{tab:nds_transfer} denote the discovery sources, while columns denote the evaluation targets. Diagonal entries represent settings in which the discovery and evaluation tasks are the same, whereas off-diagonal entries measure transfer performance without rediscovery or recalibration.

\begin{table}[t]
\centering
\caption{Cross-dataset Kendall's $\tau$ of frozen Bi-EZP proxies on NATS-Bench-TSS. Rows are discovery sources and columns are evaluation targets. Bold marks the best source for each target.}
\label{tab:tss_transfer}
\small
\setlength{\tabcolsep}{4pt}
\begin{tabular}{lccc}
\toprule
\textbf{Source $\backslash$ Target} & \textbf{C-10} & \textbf{C-100} & \textbf{IN-16} \\
\midrule
CIFAR-10       & \textbf{0.809} & 0.786 & 0.731 \\
CIFAR-100      & 0.804 & \textbf{0.791} & 0.748 \\
ImageNet-16-120& 0.799 & 0.784 & \textbf{0.775} \\
\bottomrule
\end{tabular}
\end{table}

Table~\ref{tab:tss_transfer} shows a consistent cross-dataset pattern. Each target achieves its highest correlation when the source dataset matches the target, but every off-diagonal transfer remains between $0.731$ and $0.804$. Averaged over the three matched settings, $\tau$ is $0.792$; the mean over the six transferred settings is $0.775$, a decrease of $0.017$. Transfer is strongest between CIFAR-10 and CIFAR-100, where the two directions yield $0.804$ and $0.786$. Transfer to ImageNet-16-120 is more sensitive to the source: the CIFAR-100 proxy reaches $0.748$, compared with $0.731$ for the CIFAR-10 proxy. These results indicate that the frozen program retains most of its ranking fidelity across datasets within the same topology search space, while the remaining gap reflects a measurable source--target effect.

\begin{table*}[t]
\centering
\caption{Cross-search-space Kendall's $\tau$ of frozen Bi-EZP proxies on NDS. Rows are discovery sources and columns are evaluation targets. Bold marks the best source for each target.}
\label{tab:nds_transfer}
\small
\setlength{\tabcolsep}{7pt}
\begin{tabular}{lccccc}
\toprule
\textbf{Source $\backslash$ Target} & \textbf{DARTS} & \textbf{ENAS} & \textbf{PNAS} & \textbf{NASNet} & \textbf{Amoeba} \\
\midrule
DARTS  & 0.621 & 0.558 & 0.519 & 0.486 & 0.476 \\
ENAS   & 0.625 & \textbf{0.582} & 0.548 & 0.488 & 0.429 \\
PNAS   & \textbf{0.627} & 0.563 & \textbf{0.551} & 0.485 & 0.434 \\
NASNet & 0.528 & 0.501 & 0.479 & \textbf{0.519} & 0.479 \\
Amoeba & 0.549 & 0.554 & 0.519 & 0.454 & \textbf{0.511} \\
\bottomrule
\end{tabular}
\end{table*}

The NDS matrix in Table~\ref{tab:nds_transfer} covers a larger structural shift. Off-diagonal correlations range from $0.429$ to $0.627$, with a mean of $0.515$ compared with $0.557$ on the diagonal. Transfer among DARTS, ENAS, and PNAS is particularly stable: their cross-space values fall between $0.519$ and $0.627$. The PNAS-discovered proxy attains the highest DARTS correlation ($0.627$), slightly exceeding the DARTS-discovered proxy ($0.621$), which shows that a matched source is not uniformly necessary for the best target ranking. NASNet and Amoeba expose a larger shift: proxies transferred into these targets range from $0.454$ to $0.519$ and from $0.429$ to $0.511$, respectively. These results provide evidence of cross-space reuse, while the source--target differences indicate that the proxy is not invariant to architectural-family shifts. The observed robustness may be facilitated by the separation of structural search and numerical calibration.

\subsubsection{Effectiveness and Efficiency in Open-Domain Architecture Search}

While rank correlation provides an indirect measure of proxy fidelity, the ultimate criterion of a proxy's practical value lies in its ability to guide end-to-end architecture search. To this end, we integrate Bi-EZP into a full neural architecture search pipeline within the continuous relaxation of the DARTS search space, thereby evaluating its effectiveness under realistic search dynamics.

As summarized in Table~\ref{table:cifar10}, architectures discovered under the guidance of Bi-EZP reach test errors of $2.47\%$ on CIFAR-10 and $16.10\%$ on CIFAR-100. ECP reports $2.52\%$ on CIFAR-10 under the same tabulated search cost. These results show that the rank-correlation improvements are accompanied by competitive downstream architecture quality under the reported protocol.

Bi-EZP-guided search is also compared with reinforcement-learning approaches such as ENAS and gradient-based methods including PC-DARTS and GDAS. The reported $0.06$ GPU-days correspond only to the downstream DARTS architecture-search stage and do not include the one-time offline Bi-EZP proxy-discovery cost, which comprises LLM inference, base-proxy extraction, CMA-ES evaluations, and validation. The tabulated number should therefore be interpreted as downstream search cost rather than total method-development cost.

\paragraph{Scalability to Large-Scale Datasets.}

To further evaluate the scalability of the discovered proxy beyond moderate-scale benchmarks, we extend the search process to the high-resolution ImageNet dataset. Following the established evaluation protocol in prior evolutionary proxy frameworks such as ECP~\cite{huang2025evolving}, directly evolving task-specific proxy coefficients on ImageNet is computationally prohibitive due to the substantial evaluation overhead. Therefore, to ensure both fairness and computational feasibility, we transfer the proxy structure evolved on the DARTS-CIFAR-10 search space to the DARTS-ImageNet setting, leveraging the structural similarity between the two search domains.

Under this transfer-based protocol, the architecture discovered under Bi-EZP guidance achieves a Top-1 test error of $24.7\%$ with $5.3$M parameters, as reported in Table~\ref{table:imgaenet}. The corresponding downstream architecture-search stage costs $0.07$ GPU-days; as above, this value excludes the one-time offline proxy-discovery cost. The result shows that a proxy discovered in the DARTS-CIFAR-10 setting can be used to guide the reported DARTS-ImageNet search, while broader transfer claims remain outside the evaluated setting.
\setlength{\tabcolsep}{4pt}
\begin{table*}[!htb]
\centering
% \scalebox{0.57}{
\footnotesize
\caption{Comparison of Bi-EZP with representative NAS methods on CIFAR-10 and CIFAR-100.}
\begin{tabular}{lcccccccc}

\hline
\hline
\multirow{2}{*}{\footnotesize{Architecture}}  & \multicolumn{2}{c}{Test Error (\%)} & Params    & Search Cost &\multirow{2}{*}{\footnotesize{Search Strategy}}\\ \cline{2-3}
                        & CIFAR-10         & CIFAR-100        & (M)       & (GPU-Days)  \\ \hline
DenseNet \cite{Densenet}                          & 3.46                          & 17.18                          & 25.6                        & -   & -                              \\
ResNet \cite{resnet}                              & 4.61                          & 22.1                           & 1.7                         & -   & -                              \\
VGG \cite{vggnet}                          & 6.66                          & 28.05                          & 20.1                         & - & -                               \\ \hline
ENAS  \cite{pham2018efficient}                     & 2.89                          & 19.43                              & 4.6                         & 0.5                & RL                    \\
MetaQNN \cite{metaqnn}                     & 6.92                         & 17.14                             & -                        & 100                         & RL          \\ \hline
ADARTS \cite{xue2022partial} & 2.46                          & 17.03                             & 2.9                         & 0.2  & GD  \\
$\beta$-DARTS \cite{ye2022b}                         & 2.53                        & 16.24                                          & 3.78       & 0.4     & GD        \\
PC-DARTS \cite{xu2019pc}                          & 2.57                          & 16.90                          & 3.6                         & 0.1     & GD     \\
DARTS-  \cite{chudarts} & 2.59                          & 17.51                          & 3.4                         & 0.4  & GD                                   \\
FairDARTS  \cite{fairDARTS}                  & 2.54             & 17.61           & 2.8     & 0.4     & GD        \\
SNAS \cite{SNAS} & 3.10 & 20.09 & 2.8 & 1.5& GD     \\
IS-DARTS \cite{he2024darts} & 2.56                          & -                              & 4.25                         & 0.42   & GD  \\
DARTS(2st) \cite{liu2018darts}                      & 2.76                          & -                              & 3.3                         & 1.0                 & GD                    \\
GDAS \cite{GDAS} & 2.93 & 19.18 & 3.4 & 0.2& GD     \\
DrNAS \cite{drnas} & 2.54                          & -                              & 4.0                         & 0.4   & GD            \\
DARTS+PT \cite{DARTS+PT} & 2.61 &- & 3.0 & 0.8& GD     \\
DARTS(1st) \cite{liu2018darts}                      & 3.00                          & 17.54                          & 3.4                         & 0.4          & GD                        \\ \hline
Cars \cite{CARS} & 2.62                         & -                              & 3.6                        & 0.4                           & EA         \\
GENAS \cite{xue2023neural}                          & 2.49                          & 16.96                        & 3.20                      & 0.26 & EA \\
CNN-GA (CIFAR100) \cite{sun2020automatically}   & -                        & 20.53                          & 4.1                        & 40        & EA                             \\
EAEPSO \cite{yuan2023particle}                          & 2.74                          & 16.94                          & 2.94          & 2.2               &  EA                                 \\
AmoebaNet-A \cite{real2019regularized}                    & 3.34                          & 17.63                          & 3.3                         & 3150            & EA                          \\
SLE-NAS-B \cite{huang2023split}                          & 3.47                          & 18.07                          & 0.94                       & 0.4     & EA \\
NPENAS-NP ~\cite{NPENAS} & 3.62                        & 26.76                             & 3.5                      & 1.8       & EA                             \\
SaDENAS \cite{han2024sadenas}                          & 2.59                          & 16.91                         & 3.24                      & 0.2      & EA \\
CNN-GA (CIFAR10) \cite{sun2020automatically}    & 3.22                          & -                          & 2.9                         & 35   & EA                                   \\
EPCNAS-C \cite{huang2022particle}                           & 3.07                          & 18.36                          & 1.16    & 1.10                     & EA                                 \\
NSGANetV1-A2 \cite{lu2020multiobjective} & 2.65                          & 17.42                              & \textbf{0.9}                      & 27                           & EA            \\ \hline
NTK \cite{chen2021neural}                                 & 2.89                          & 20.30                             & 4.1                       & 0.21     & ZS \\
GraSP \cite{wang2020picking}                             & 2.73                          & 22.65                              & 3.3                       & 0.1      & ZS \\
MeCo \cite{jiang2023meco}                               & 2.69                          & 16.86                             & 4.2                       & 0.08     & ZS \\
SynFlow \cite{tanaka2020pruning}                         & 2.96                          & 19.82                              & 5.1                       & 0.03     & ZS \\
NASWOT \cite{abdelfattah2021zero}                           & 2.77                          & 22.90                              & 4.8                       & 0.06     & ZS \\
Sweetimator \cite{yang2023sweet}                 & 2.54                          & -                              & 4.6                       & 0.05     & ZS \\
ZiCo \cite{li2023zico}                               & 2.80                          & 19.54                             & 5.1                       & 0.04     & ZS \\
SNIP \cite{lee2018snip}                               & 2.90                          & 19.95                              & 4.0                       & 0.04     & ZS \\ 
ECP~\cite{huang2025evolving}                                        & 2.52                          & -                              & 4.5                       & \textbf{0.06}     & ZS \\ \hline
Bi-EZP        & \textbf{2.47}                       & \textbf{16.10}                        & 3.3                        & \textbf{0.06}             & ZS            \\ \hline \hline

\end{tabular}

\label{table:cifar10}
\end{table*}

\begin{table}[!htb]
\centering
% \captionsetup{font={footnotesize}}
% \scalebox{1.0}{
\caption{Comparison with representative NAS methods on ImageNet.}
\begin{threeparttable}
% \scalebox{0.7}{
\footnotesize
\begin{tabular}{lcccc}
\hline
\hline
\scriptsize
\multirow{2}{*}{\footnotesize Architecture} & Test error & Search cost & Params & \multirow{2}{*}{\footnotesize Search Strategy} \\
 & top-1(\%) & (GPU-Days) & (M) & \\ \hline
ResNet~\cite{resnet}& 30.1 & - & 6.6 & Manual \\
VGG~\cite{vggnet} & 29.4 & - & 4.2 & Manual \\
DenseNet~\cite{Densenet} & 29.4 & - & 4.2 & Manual \\ \hline
DARTS(2st)~\cite{liu2018darts}& 26.7 & 1.0 & 4.7 & GD \\
SNAS~\cite{SNAS} & 27.3 & 1.5 & \textbf{2.8} & GD \\
PC-DARTS~\cite{xu2019pc} & 25.1 & 0.1 & 4.7  & GD \\ 
GDAS~\cite{GDAS} & 26.0 & 0.3 & 3.4 & GD \\
ProxylessNAS~\cite{cai2018proxylessnas} & 24.9  & 8.3    & 7.1   & GD \\ 
DARTS+PT~\cite{DARTS+PT} &25.5&3.4&4.7 & GD \\ \hline
AmoebaNet-B~\cite{real2019regularized} & 26.0 & 3150 & 5.3 & EA \\
CARS~\cite{CARS} & 24.8 & 0.4 & 5.1 & EA \\
NSGANetV1-A2~\cite{lu2020multiobjective} & 25.5 & 27 & 4.1 & EA \\ 
EAEPSO~\cite{yuan2023particle} & 26.9 & 4.0 & 4.9 & EA \\
EPCNAS-C2~\cite{huang2022particle} & 27.1 & 1.17 & 3.0 & EA \\ 
EG-NAS~\cite{cai2024eg} & 24.9 & 0.1 & 5.3 & EA \\
LAPT-NAS~\cite{zhou2025design} & 24.9 & 2 & 4.6 & EA \\ \hline
TE-NAS~\cite{chen2021neural}& 26.2 & \textbf{0.05} & 6.3 & ZS \\
NASI-ADA~\cite{shu2021nasi} & 24.8 & 0.07 & 5.2 & ZS \\
QE-NAS~\cite{sun2022entropy}& 25.5 & 0.02 & 3.2 & ZS\\
ECP~\cite{huang2025evolving} & \textbf{24.7} & 0.07 & 6.7 & ZS \\
\hline
Bi-EZP & \textbf{24.7} & 0.07 & 5.3 & ZS  \\
\hline \hline
\end{tabular}
\end{threeparttable}
\label{table:imgaenet}
\end{table}

\begin{table*}[htbp]
\centering
\footnotesize 
\caption{Ablation study of different components in Bi-EZP on NATS-Bench.}
\label{tab:ablation}

\begin{tabular}{lcccccc}
\toprule
\multirow{2}{*}{\textbf{Method}} 
& \multicolumn{3}{c}{\textbf{NATS-Bench-TSS}} 
& \multicolumn{3}{c}{\textbf{NATS-Bench-SSS}} \\

\cmidrule(lr){2-4} \cmidrule(lr){5-7}

& \textbf{C-10} & \textbf{C-100} & \textbf{IN-16} 
& \textbf{C-10} & \textbf{C-100} & \textbf{IN-16} \\

\midrule

w/o Bilevel Structure 
& 0.783 & 0.741 & 0.731 
& 0.743 & 0.678 & 0.806 \\

w/o LLM (ECP Aggregation + CMA-ES) 
& 0.784 & 0.771 & 0.740 
& 0.769 & 0.656 & 0.805 \\

Bi-EZP (GA) 
& 0.807 & 0.783 & 0.737 
& 0.751 & 0.725 & 0.580 \\

Bi-EZP (DE) 
& 0.788 & 0.771 & 0.752 
& 0.768 & 0.713 & 0.805 \\

Bi-EZP (PSO) 
& 0.789 & 0.774 & 0.756 
& 0.780 & 0.719 & 0.798 \\

\midrule

\textbf{Bi-EZP} 
& \textbf{0.809} & \textbf{0.791} & \textbf{0.775} 
& \textbf{0.792} & \textbf{0.733} & \textbf{0.826} \\

\bottomrule
\end{tabular}
\end{table*}

\begin{figure}[t]
  \centering
  % 左侧图
  \begin{minipage}{0.48\textwidth}
    \centering
    \includegraphics[width=\linewidth]{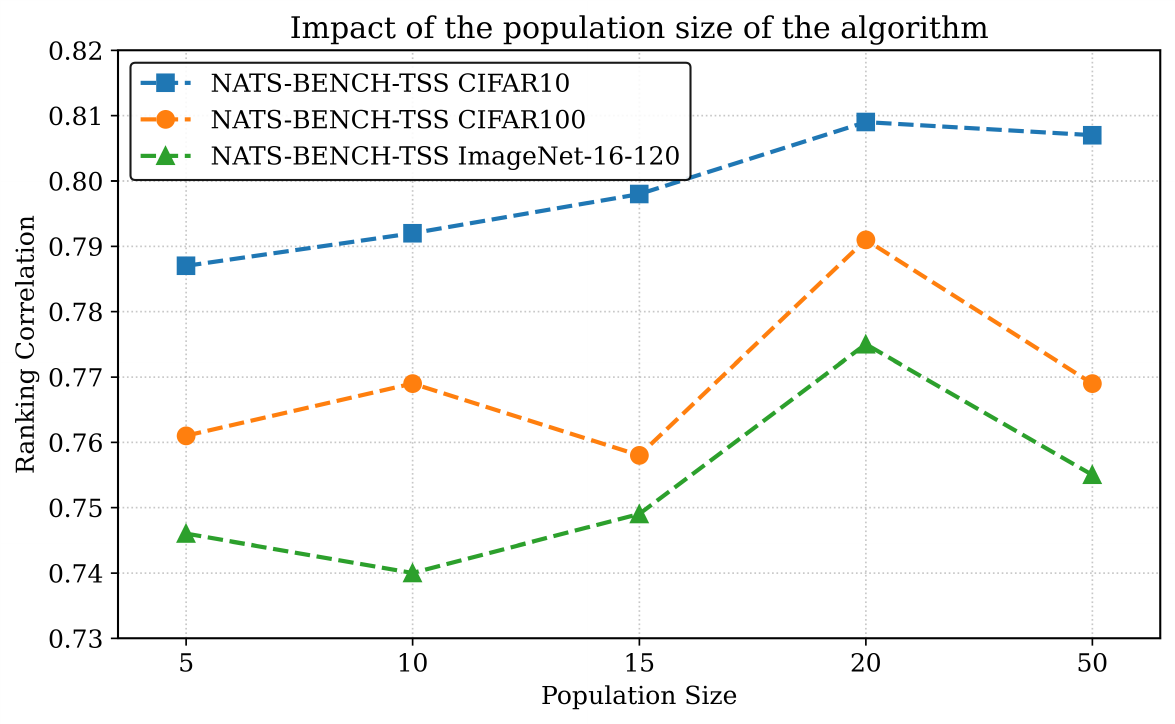}
\caption{Sensitivity of Bi-EZP to population size on NATS-Bench-TSS.}
    \label{ab_pop_size}
  \end{minipage}\hfill 
  \begin{minipage}{0.48\textwidth}
    \centering
    \includegraphics[width=\linewidth]{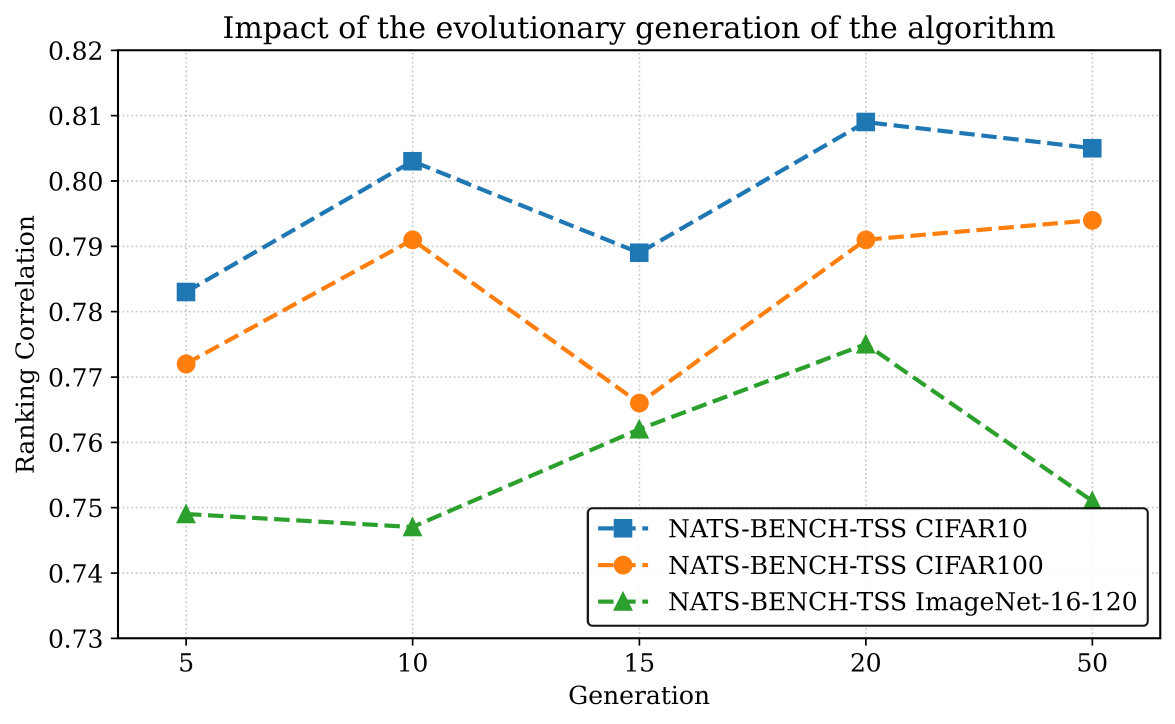}
\caption{Sensitivity of Bi-EZP to the number of evolutionary generations on NATS-Bench-TSS.}
    \label{ab_gen}
  \end{minipage}
\end{figure}
% \begin{figure}[t]
%   \centering 
%     \includegraphics[width=0.48 \textwidth]{pop_size_ab.pdf}
%     % \captionsetup{font={footnotesize}}
%   \caption{The Impact of population size on the proposed framework.}
%   \label{ab_pop_size}
% \end{figure}

% \begin{figure}[t]
%   \centering 
%     \includegraphics[width=0.48 \textwidth]{gen_ab.pdf}
%     % \captionsetup{font={footnotesize}}
%   \caption{The Impact of evolutionary generation on the proposed framework.}
%   \label{ab_gen}
% \end{figure}

\begin{figure}[t]
  \centering 
    \includegraphics[width=0.48 \textwidth]{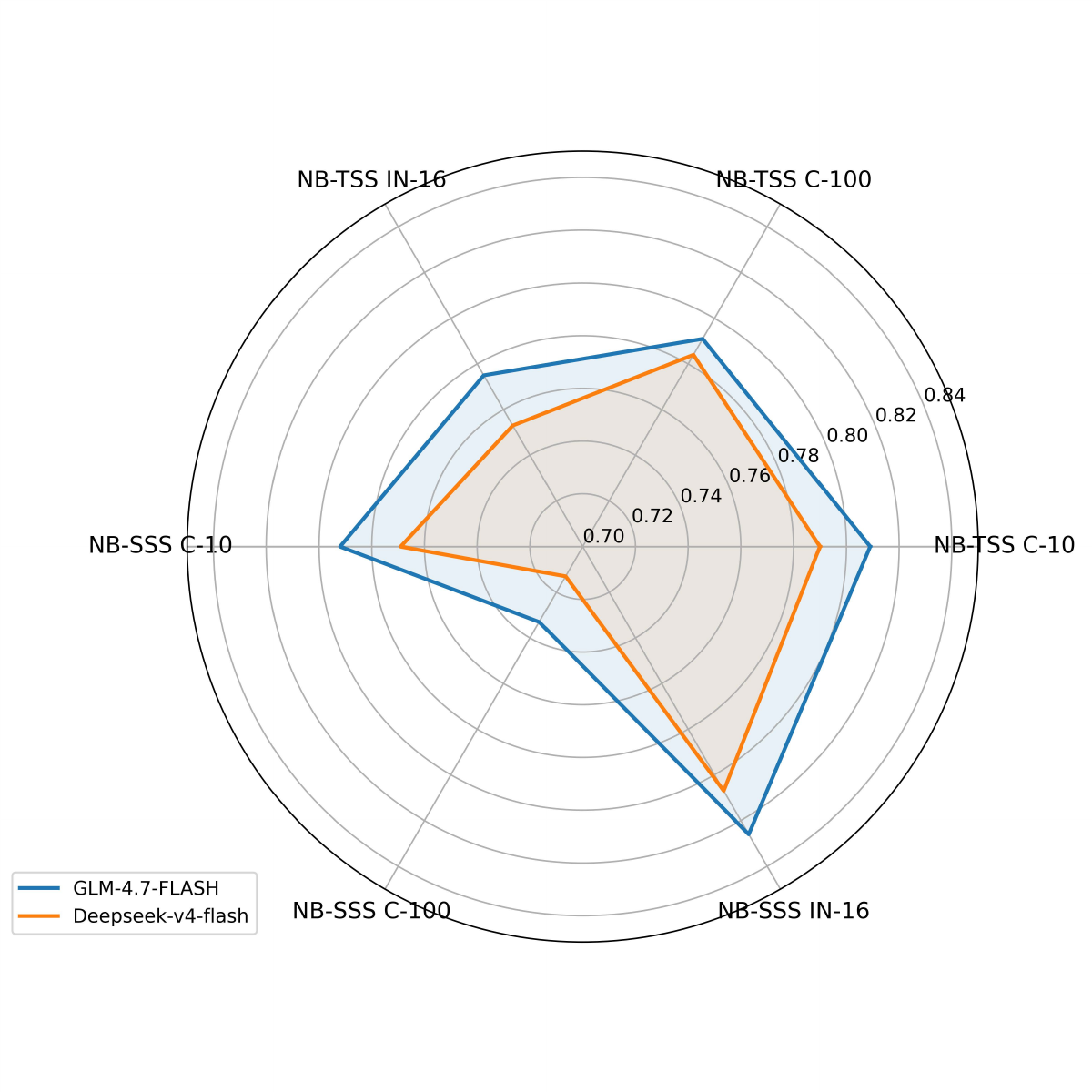}
    % \captionsetup{font={footnotesize}}
  \caption{Performance Comparison under Different LLMs of the proposed framework.}
  \label{fig:radar_chart}
\end{figure}

\subsection{Ablation Study and Hyperparameter Analysis}
To understand the contribution of the main mechanisms in Bi-EZP, we conduct an ablation study on NATS-Bench (Table~\ref{tab:ablation}) and analyze sensitivity to population size and the number of evolutionary generations (Figs.~\ref{ab_pop_size} and~\ref{ab_gen}). The updated sweeps evaluate both quantities up to 50, allowing us to distinguish a useful operating region from a simple monotonic budget effect.

\subsubsection{The Efficacy of Bilevel Decoupling and LLM-Driven Semantic Search}
Bi-EZP uses a bilevel design to separate discrete structure proposal from continuous parameter calibration. To assess this separation, we first evaluate a variant denoted as \textbf{``w/o Bilevel Structure"}. In this configuration, the LLM directly generates both the symbolic computation graph and its numerical coefficients, collapsing the search into a single level. Table~\ref{tab:ablation} reports lower correlation for this variant across the evaluated datasets, including a reduction from 0.809 to 0.783 on TSS CIFAR-10. These observations are consistent with a benefit from calibrating continuous coefficients with a dedicated lower-level optimizer under the reported protocol.

Secondly, to assess the effect of the upper-level proposal mechanism, we test the \textbf{``w/o LLM (ECP Aggregation + CMA-ES)"} variant. This baseline replaces the LLM proposal operator with the ECP aggregation structure while retaining CMA-ES for lower-level calibration. It is inferior to the complete framework under the reported protocol (e.g., 0.784 versus 0.809 on TSS CIFAR-10). Following the motivation of EoH~\cite{liu2024evolution}, we use an LLM because it can express candidate heuristics jointly as a short functional description and executable code, and can propose context-conditioned transformations without requiring us to enumerate a closed grammar of all admissible compositions. Selection remains empirical: generated programs are retained only after interface validation, CMA-ES calibration, and validation-set rank-correlation evaluation. Since we do not report matched expression-size or search-efficiency measurements for a GP system, we draw no comparative conclusion about expression size or interpretability.

\subsubsection{Comparison of Lower-Level Optimizers}
Once the upper-level LLM formulates the explicit parameter boundaries, the lower-level solver must navigate this continuous manifold to maximize the Kendall rank correlation strictly on the inner-training set. Table~\ref{tab:ablation} presents a comparative analysis of different continuous optimizers, replacing CMA-ES with Genetic Algorithm (GA), Differential Evolution (DE), and Particle Swarm Optimization (PSO). 

Under the reported settings, \textbf{Bi-EZP (CMA-ES)} obtains higher rank correlation than the GA-, DE-, and PSO-based variants across the evaluated search spaces. We use CMA-ES because its derivative-free covariance adaptation is compatible with the non-differentiable rank-correlation objective and can model dependencies among continuous parameters. This ablation is limited to the three alternative optimizers in Table~\ref{tab:ablation} and does not establish that CMA-ES is optimal among all possible lower-level solvers.

\subsubsection{Impact of Evolutionary Hyperparameters}
We evaluate population sizes and generation budgets in $\{5,10,15,20,50\}$, as shown in Figs.~\ref{ab_pop_size} and~\ref{ab_gen}. Each point reports the final Kendall correlation for one configuration; no uncertainty interval is available, so the analysis concerns the observed sensitivity rather than statistical monotonicity.

For population size, performance generally improves from $P=5$ to $P=20$. On CIFAR-10, $\tau$ rises from approximately $0.787$ to $0.809$; CIFAR-100 improves from $0.761$ to $0.791$; and ImageNet-16-120 improves from $0.746$ to $0.775$, despite smaller non-monotonic changes at intermediate sizes. Expanding the population to $P=50$ does not yield a further gain: CIFAR-10 remains close at about $0.807$, while CIFAR-100 and ImageNet-16-120 fall to about $0.769$ and $0.755$. Thus, additional candidates beyond 20 increase the number of LLM generations and CMA-ES calibrations without improving the observed aggregate outcome. We use $P=20$ as the best tested balance across the three tasks.

The generation sweep is also non-monotonic. Increasing the budget from 5 to 20 raises the observed correlation from about $0.783/0.772/0.749$ to $0.809/0.791/0.775$ on CIFAR-10, CIFAR-100, and ImageNet-16-120, respectively, with a common dip at 15 generations. Extending the run to 50 generations produces mixed results: CIFAR-100 increases slightly to about $0.794$, whereas CIFAR-10 declines to about $0.805$ and ImageNet-16-120 to about $0.751$. The extra generations therefore do not provide a consistent cross-dataset benefit. We retain $T_{max}=20$ because it gives the strongest balanced result among the tested budgets and requires less than half the candidate evaluations of the 50-generation setting. These curves support a practical budget choice, but they do not by themselves establish convergence or identify the causes of the intermediate fluctuations.

\subsubsection{Performance Sensitivity under Different LLMs}
To examine sensitivity to the upper-level generative backbone, we compare GLM-4.7-Flash~\cite{glm2024chatglm} and DeepSeek-V4-Flash~\cite{deepseek2026deepseek}. Figure~4 reports competitive rank correlations for both variants across the evaluated benchmarks, with GLM-4.7-Flash producing higher correlations in several settings, including ImageNet-16-120 under TSS and SSS. This two-model comparison shows that the evaluated procedure can operate with either tested backbone, but it does not establish invariance across LLM providers, model scales, or decoding configurations.

\begin{figure*}[t]
  \centering 
    \includegraphics[width=0.85\textwidth]{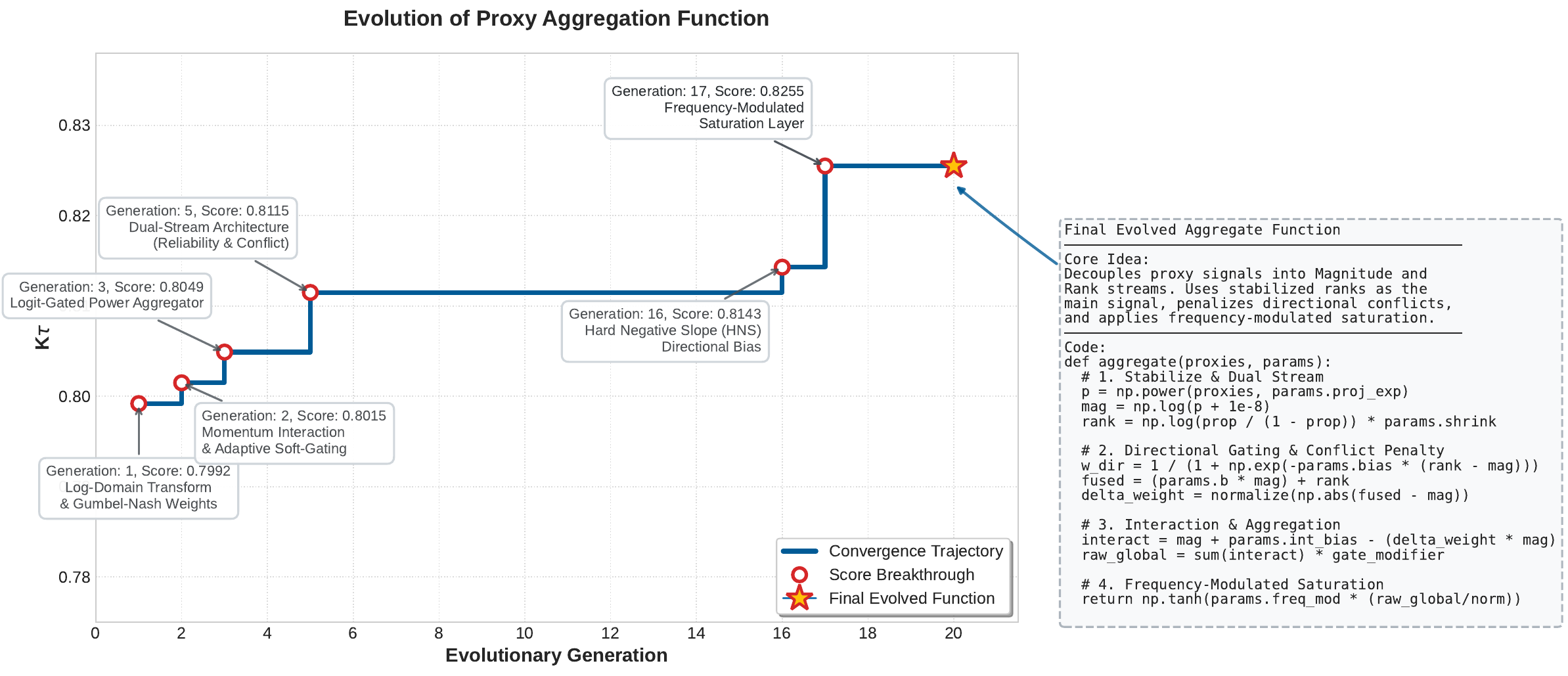}
    % \captionsetup{font={footnotesize}}
  \caption{The evolutionary trajectory of the proposed framework over 20 generations on CIFAR10 of the NATS-Bench TSS benchmark.}
  \label{fig:evo_tra}
\end{figure*}

\subsection{Evolutionary Trajectory and Search Dynamics}
\label{sec:trajectory_analysis}
To elucidate the search dynamics of the proposed framework, we analyze the step-wise convergence of the Kendall rank correlation coefficient ($K_\tau$) over 20 generations (Figure~\ref{fig:evo_tra}). 

The trajectory shows a rapid increase in the best observed validation correlation during Generations 1-5 ($K_\tau$ reaching 0.8115), followed by a plateau during Generations 5-16 and a later improvement. This trace describes the best-so-far fitness in one run; it does not reveal why individual proposals were rejected and should not be interpreted as evidence of global convergence. The best-so-far correlation reaches $0.8255$ at Generation 17 and remains unchanged through the recorded budget.

\section{CONCLUSION}
Bi-EZP separates LLM-based program proposal from CMA-ES-based continuous parameter calibration for automated ensemble zero-shot proxy discovery. Under the evaluated protocols, the resulting proxies achieve higher rank correlation than the compared baselines on NATS-Bench and NDS and can be used as evaluation signals in downstream DARTS search.

Despite its efficacy, the current framework exhibits certain limitations. The offline proxy discovery phase introduces non-trivial token costs and latency due to iterative LLM querying. Additionally, the quality of the discovered proxy is bounded by the intrinsic properties of the pre-selected base metrics, and the structural diversity remains sensitive to the contextual alignment of prompt constraints. Future work will focus on expanding the bilevel formulation into a multi-objective paradigm to handle hardware-aware constraints like device latency.

\bibliographystyle{IEEEtran}
\bibliography{ref}

@article{he2021automl,
  title={AutoML: A survey of the state-of-the-art},
  author={He, Xin and Zhao, Kaiyong and Chu, Xiaowen},
  journal={Knowledge-based systems},
  volume={212},
  pages={106622},
  year={2021},
  publisher={Elsevier}
}

@article{vggnet,
  title={Very deep convolutional networks for large-scale image recognition},
  author={Simonyan, Karen and Zisserman, Andrew},
  journal={arXiv preprint arXiv:1409.1556},
  year={2014}
}

@inproceedings{resnet,
  title={Deep residual learning for image recognition},
  author={He, Kaiming and Zhang, Xiangyu and Ren, Shaoqing and Sun, Jian},
  booktitle={Proceedings of the IEEE conference on computer vision and pattern recognition},
  pages={770--778},
  year={2016}
}

@article{zoph2016neural,
  title={Neural architecture search with reinforcement learning},
  author={Zoph, B},
  journal={arXiv preprint arXiv:1611.01578},
  year={2016}
}

@inproceedings{pham2018efficient,
  title={Efficient neural architecture search via parameters sharing},
  author={Pham, Hieu and Guan, Melody and Zoph, Barret and Le, Quoc and Dean, Jeff},
  booktitle={International conference on machine learning},
  pages={4095--4104},
  year={2018},
  organization={PMLR}
}

@inproceedings{real2017large,
  title={Large-scale evolution of image classifiers},
  author={Real, Esteban and Moore, Sherry and Selle, Andrew and Saxena, Saurabh and Suematsu, Yutaka Leon and Tan, Jie and Le, Quoc V and Kurakin, Alexey},
  booktitle={International conference on machine learning},
  pages={2902--2911},
  year={2017},
  organization={PMLR}
}

@article{sun2020automatically,
  title={Automatically designing CNN architectures using the genetic algorithm for image classification},
  author={Sun, Yanan and Xue, Bing and Zhang, Mengjie and Yen, Gary G and Lv, Jiancheng},
  journal={IEEE transactions on cybernetics},
  volume={50},
  number={9},
  pages={3840--3854},
  year={2020},
  publisher={IEEE}
}

@inproceedings{cai2024eg,
  title={EG-NAS: Neural architecture search with fast evolutionary exploration},
  author={Cai, Zicheng and Chen, Lei and Liu, Peng and Ling, Tongtao and Lai, Yutao},
  booktitle={Proceedings of the AAAI conference on artificial intelligence},
  volume={38},
  number={10},
  pages={11159--11167},
  year={2024}
}

@article{liu2018darts,
  title={Darts: Differentiable architecture search},
  author={Liu, Hanxiao and Simonyan, Karen and Yang, Yiming},
  journal={arXiv preprint arXiv:1806.09055},
  year={2018}
}

@article{xue2022partial,
  title={Partial connection based on channel attention for differentiable neural architecture search},
  author={Xue, Yu and Qin, Jiafeng},
  journal={IEEE Transactions on Industrial Informatics},
  volume={19},
  number={5},
  pages={6804--6813},
  year={2022},
  publisher={IEEE}
}

@inproceedings{ye2022b,
  title={b-darts: Beta-decay regularization for differentiable architecture search},
  author={Ye, Peng and Li, Baopu and Li, Yikang and Chen, Tao and Fan, Jiayuan and Ouyang, Wanli},
  booktitle={proceedings of the IEEE/CVF conference on computer vision and pattern recognition},
  pages={10874--10883},
  year={2022}
}

@article{li2024zero,
  title={Zero-shot neural architecture search: Challenges, solutions, and opportunities},
  author={Li, Guihong and Hoang, Duc and Bhardwaj, Kartikeya and Lin, Ming and Wang, Zhangyang and Marculescu, Radu},
  journal={IEEE Transactions on Pattern Analysis and Machine Intelligence},
  volume={46},
  number={12},
  pages={7618--7635},
  year={2024},
  publisher={IEEE}
}

@article{lee2018snip,
  title={Snip: Single-shot network pruning based on connection sensitivity},
  author={Lee, Namhoon and Ajanthan, Thalaiyasingam and Torr, Philip HS},
  journal={arXiv preprint arXiv:1810.02340},
  year={2018}
}

@article{wang2020picking,
  title={Picking winning tickets before training by preserving gradient flow},
  author={Wang, Chaoqi and Zhang, Guodong and Grosse, Roger},
  journal={arXiv preprint arXiv:2002.07376},
  year={2020}
}

@article{abdelfattah2021zero,
  title={Zero-cost proxies for lightweight NAS},
  author={Abdelfattah, Mohamed S and Mehrotra, Abhinav and Dudziak, {\L}ukasz and Lane, Nicholas D},
  journal={arXiv preprint arXiv:2101.08134},
  year={2021}
}

@article{chen2021neural,
  title={Neural architecture search on imagenet in four gpu hours: A theoretically inspired perspective},
  author={Chen, Wuyang and Gong, Xinyu and Wang, Zhangyang},
  journal={arXiv preprint arXiv:2102.11535},
  year={2021}
}

@inproceedings{mellor2021neural,
  title={Neural architecture search without training},
  author={Mellor, Joe and Turner, Jack and Storkey, Amos and Crowley, Elliot J},
  booktitle={International conference on machine learning},
  pages={7588--7598},
  year={2021},
  organization={PMLR}
}

@article{ning2021evaluating,
  title={Evaluating efficient performance estimators of neural architectures},
  author={Ning, Xuefei and Tang, Changcheng and Li, Wenshuo and Zhou, Zixuan and Liang, Shuang and Yang, Huazhong and Wang, Yu},
  journal={Advances in Neural Information Processing Systems},
  volume={34},
  pages={12265--12277},
  year={2021}
}

@inproceedings{lin2021zen,
  title={Zen-nas: A zero-shot nas for high-performance image recognition},
  author={Lin, Ming and Wang, Pichao and Sun, Zhenhong and Chen, Hesen and Sun, Xiuyu and Qian, Qi and Li, Hao and Jin, Rong},
  booktitle={Proceedings of the IEEE/CVF international conference on computer vision},
  pages={347--356},
  year={2021}
}

@article{jiang2023meco,
  title={Meco: zero-shot nas with one data and single forward pass via minimum eigenvalue of correlation},
  author={Jiang, Tangyu and Wang, Haodi and Bie, Rongfang},
  journal={Advances in Neural Information Processing Systems},
  volume={36},
  pages={61020--61047},
  year={2023}
}

@article{li2023zico,
  title={Zico: Zero-shot nas via inverse coefficient of variation on gradients},
  author={Li, Guihong and Yang, Yuedong and Bhardwaj, Kartikeya and Marculescu, Radu},
  journal={arXiv preprint arXiv:2301.11300},
  year={2023}
}

@article{peng2024swap,
  title={SWAP-NAS: Sample-wise activation patterns for ultra-fast NAS},
  author={Peng, Yameng and Song, Andy and Fayek, Haytham M and Ciesielski, Vic and Chang, Xiaojun},
  journal={arXiv preprint arXiv:2403.04161},
  year={2024}
}

@inproceedings{casarin2025swag,
  title={L-swag: Layer-sample wise activation with gradients information for zero-shot nas on vision transformers},
  author={Casarin, Sofia and Escalera, Sergio and Lanz, Oswald},
  booktitle={Proceedings of the IEEE/CVF Conference on Computer Vision and Pattern Recognition},
  pages={4441--4451},
  year={2025}
}

@inproceedings{wei2024auto,
  title={Auto-prox: Training-free vision transformer architecture search via automatic proxy discovery},
  author={Wei, Zimian and Dong, Peijie and Hui, Zheng and Li, Anggeng and Li, Lujun and Lu, Menglong and Pan, Hengyue and Li, Dongsheng},
  booktitle={Proceedings of the aaai conference on artificial intelligence},
  volume={38},
  number={14},
  pages={15814--15822},
  year={2024}
}

@inproceedings{lee2024az,
  title={Az-nas: Assembling zero-cost proxies for network architecture search},
  author={Lee, Junghyup and Ham, Bumsub},
  booktitle={Proceedings of the IEEE/CVF conference on computer vision and pattern recognition},
  pages={5893--5903},
  year={2024}
}

@article{akhauri2022eznas,
  title={EZNAS: evolving zero-cost proxies for neural architecture scoring},
  author={Akhauri, Yash and Munoz, Juan and Jain, Nilesh and Iyer, Ravishankar},
  journal={Advances in Neural Information Processing Systems},
  volume={35},
  pages={30459--30470},
  year={2022}
}

@inproceedings{dong2025parzc,
  title={Parzc: Parametric zero-cost proxies for efficient nas},
  author={Dong, Peijie and Li, Lujun and Tang, Zhenheng and Liu, Xiang and Wei, Zimian and Wang, Qiang and Chu, Xiaowen},
  booktitle={Proceedings of the AAAI Conference on Artificial Intelligence},
  volume={39},
  number={15},
  pages={16327--16335},
  year={2025}
}

@inproceedings{linper,
  title={Per-Architecture Training-Free Metric Optimization for Neural Architecture Search},
  author={Lin, Mingzhuo and Luo, Jianping},
  booktitle={The Thirty-ninth Annual Conference on Neural Information Processing Systems}
}

@article{phan2025hand,
  title={From hand-crafted metrics to evolved training-free performance predictors for neural architecture search via symbolic regression},
  author={Phan, Quan Minh and Luong, Ngoc Hoang},
  journal={Neurocomputing},
  pages={132440},
  year={2025},
  publisher={Elsevier}
}

@inproceedings{huang2025evolving,
  title={Evolving Comprehensive Proxies for Zero-Shot Neural Architecture Search},
  author={Huang, Junhao and Xue, Bing and Sun, Yanan and Zhang, Mengjie},
  booktitle={Proceedings of the Genetic and Evolutionary Computation Conference},
  pages={1246--1254},
  year={2025}
}

@article{chen2023evoprompting,
  title={Evoprompting: Language models for code-level neural architecture search},
  author={Chen, Angelica and Dohan, David and So, David},
  journal={Advances in neural information processing systems},
  volume={36},
  pages={7787--7817},
  year={2023}
}

@article{zheng2023genius,
  title={Can gpt-4 perform neural architecture search?},
  author={Zheng, Mingkai and Su, Xiu and You, Shan and Wang, Fei and Qian, Chen and Xu, Chang and Albanie, Samuel},
  journal={arXiv preprint arXiv:2304.10970},
  year={2023}
}

@inproceedings{zhou2025design,
  title={Design principle transfer in neural architecture search via large language models},
  author={Zhou, Xun and Wu, Xingyu and Feng, Liang and Lu, Zhichao and Tan, Kay Chen},
  booktitle={Proceedings of the AAAI Conference on Artificial Intelligence},
  volume={39},
  number={21},
  pages={23000--23008},
  year={2025}
}

@inproceedings{nasir2024llmatic,
  title={Llmatic: neural architecture search via large language models and quality diversity optimization},
  author={Nasir, Muhammad Umair and Earle, Sam and Togelius, Julian and James, Steven and Cleghorn, Christopher},
  booktitle={proceedings of the Genetic and Evolutionary Computation Conference},
  pages={1110--1118},
  year={2024}
}

@inproceedings{ji2025rz,
  title={Rz-nas: Enhancing llm-guided neural architecture search via reflective zero-cost strategy},
  author={Ji, Zipeng and Zhu, Guanghui and Yuan, Chunfeng and Huang, Yihua},
  booktitle={Forty-second International Conference on Machine Learning},
  year={2025}
}

@article{cai2025seki,
  title={Seki: Self-evolution and knowledge inspiration based neural architecture search via large language models},
  author={Cai, Zicheng and Tang, Yaohua and Lai, Yutao and Wang, Hua and Chen, Zhi and Chen, Hao},
  journal={arXiv preprint arXiv:2502.20422},
  year={2025}
}

@article{huang2023split,
  title={Split-level evolutionary neural architecture search with elite weight inheritance},
  author={Huang, Junhao and Xue, Bing and Sun, Yanan and Zhang, Mengjie and Yen, Gary G},
  journal={IEEE transactions on neural networks and learning systems},
  volume={35},
  number={10},
  pages={13523--13537},
  year={2023},
  publisher={IEEE}
}

@article{hansen2003reducing,
  title={Reducing the time complexity of the derandomized evolution strategy with covariance matrix adaptation (CMA-ES)},
  author={Hansen, Nikolaus and M{\"u}ller, Sibylle D and Koumoutsakos, Petros},
  journal={Evolutionary computation},
  volume={11},
  number={1},
  pages={1--18},
  year={2003},
  publisher={MIT Press}
}

@article{dong2021nats,
  title={Nats-bench: Benchmarking nas algorithms for architecture topology and size},
  author={Dong, Xuanyi and Liu, Lu and Musial, Katarzyna and Gabrys, Bogdan},
  journal={IEEE transactions on pattern analysis and machine intelligence},
  volume={44},
  number={7},
  pages={3634--3646},
  year={2021},
  publisher={IEEE}
}

@inproceedings{radosavovic2019network,
  title={On network design spaces for visual recognition},
  author={Radosavovic, Ilija and Johnson, Justin and Xie, Saining and Lo, Wan-Yen and Doll{\'a}r, Piotr},
  booktitle={Proceedings of the IEEE/CVF international conference on computer vision},
  pages={1882--1890},
  year={2019}
}

@article{krizhevsky2009learning,
  title={Learning multiple layers of features from tiny images},
  author={Krizhevsky, Alex and Hinton, Geoffrey and others},
  year={2009},
  publisher={Toronto, ON, Canada}
}

@inproceedings{xiang2023zero,
  title={Zero-cost operation scoring in differentiable architecture search},
  author={Xiang, Lichuan and Dudziak, Lukasz and Abdelfattah, Mohamed S and Chau, Thomas and Lane, Nicholas D and Wen, Hongkai},
  booktitle={Proceedings of the AAAI conference on artificial intelligence},
  volume={37},
  number={9},
  pages={10453--10463},
  year={2023}
}

@article{tanaka2020pruning,
  title={Pruning neural networks without any data by iteratively conserving synaptic flow},
  author={Tanaka, Hidenori and Kunin, Daniel and Yamins, Daniel L and Ganguli, Surya},
  journal={Advances in neural information processing systems},
  volume={33},
  pages={6377--6389},
  year={2020}
}

@article{huang2025efficient,
  title={Efficient perturbation-aware distinguishing score for zero-shot neural architecture search},
  author={Huang, Junhao and Xue, Bing and Sun, Yanan and Zhang, Mengjie and Yen, Gary G},
  journal={Applied Soft Computing},
  pages={113447},
  year={2025},
  publisher={Elsevier}
}

@article{wu2023training,
  title={A training-free neural architecture search algorithm based on search economics},
  author={Wu, Meng-Ting and Lin, Hung-I and Tsai, Chun-Wei},
  journal={IEEE Transactions on Evolutionary Computation},
  volume={28},
  number={2},
  pages={445--459},
  year={2023},
  publisher={IEEE}
}

@article{zhang2021gradsign,
  title={Gradsign: Model performance inference with theoretical insights},
  author={Zhang, Zhihao and Jia, Zhihao},
  journal={arXiv preprint arXiv:2110.08616},
  year={2021}
}

@inproceedings{sun2023unleashing,
  title={Unleashing the power of gradient signal-to-noise ratio for zero-shot nas},
  author={Sun, Zihao and Sun, Yu and Yang, Longxing and Lu, Shun and Mei, Jilin and Zhao, Wenxiao and Hu, Yu},
  booktitle={Proceedings of the IEEE/CVF international conference on computer vision},
  pages={5763--5773},
  year={2023}
}

@inproceedings{Densenet,
  title={Densely connected convolutional networks},
  author={Huang, Gao and Liu, Zhuang and Van Der Maaten, Laurens and Weinberger, Kilian Q},
  booktitle={Proceedings of the IEEE conference on computer vision and pattern recognition},
  pages={4700--4708},
  year={2017}
}

@article{metaqnn,
  title={Designing neural network architectures using reinforcement learning},
  author={Baker, Bowen and Gupta, Otkrist and Naik, Nikhil and Raskar, Ramesh},
  journal={arXiv preprint arXiv:1611.02167},
  year={2016}
}

@article{xu2019pc,
  title={Pc-darts: Partial channel connections for memory-efficient architecture search},
  author={Xu, Yuhui and Xie, Lingxi and Zhang, Xiaopeng and Chen, Xin and Qi, Guo-Jun and Tian, Qi and Xiong, Hongkai},
  journal={arXiv preprint arXiv:1907.05737},
  year={2019}
}

@inproceedings{fairDARTS,
  title={Fair darts: Eliminating unfair advantages in differentiable architecture search},
  author={Chu, Xiangxiang and Zhou, Tianbao and Zhang, Bo and Li, Jixiang},
  booktitle={European conference on computer vision},
  pages={465--480},
  year={2020},
  organization={Springer}
}

@inproceedings{GDAS,
  title={Searching for a robust neural architecture in four gpu hours},
  author={Dong, Xuanyi and Yang, Yi},
  booktitle={Proceedings of the IEEE/CVF conference on computer vision and pattern recognition},
  pages={1761--1770},
  year={2019}
}

@article{SNAS,
  title={SNAS: stochastic neural architecture search},
  author={Xie, Sirui and Zheng, Hehui and Liu, Chunxiao and Lin, Liang},
  journal={arXiv preprint arXiv:1812.09926},
  year={2018}
}

@article{DARTS+PT,
  title={Rethinking architecture selection in differentiable NAS},
  author={Wang, Ruochen and Cheng, Minhao and Chen, Xiangning and Tang, Xiaocheng and Hsieh, Cho-Jui},
  journal={arXiv preprint arXiv:2108.04392},
  year={2021}
}

@inproceedings{chudarts,
  title={DARTS-: Robustly Stepping out of Performance Collapse Without Indicators},
  author={Chu, Xiangxiang and Wang, Xiaoxing and Zhang, Bo and Lu, Shun and Wei, Xiaolin and Yan, Junchi},
  booktitle={International Conference on Learning Representations}
}

@inproceedings{he2024darts,
  title={IS-DARTS: stabilizing DARTS through precise measurement on candidate importance},
  author={He, Hongyi and Liu, Longjun and Zhang, Haonan and Zheng, Nanning},
  booktitle={Proceedings of the AAAI Conference on Artificial Intelligence},
  volume={38},
  number={11},
  pages={12367--12375},
  year={2024}
}

@article{drnas,
  title={Drnas: Dirichlet neural architecture search},
  author={Chen, Xiangning and Wang, Ruochen and Cheng, Minhao and Tang, Xiaocheng and Hsieh, Cho-Jui},
  journal={arXiv preprint arXiv:2006.10355},
  year={2020}
}

@inproceedings{real2019regularized,
  title={Regularized evolution for image classifier architecture search},
  author={Real, Esteban and Aggarwal, Alok and Huang, Yanping and Le, Quoc V},
  booktitle={Proceedings of the aaai conference on artificial intelligence},
  volume={33},
  number={01},
  pages={4780--4789},
  year={2019}
}

@article{lu2020multiobjective,
  title={Multiobjective evolutionary design of deep convolutional neural networks for image classification},
  author={Lu, Zhichao and Whalen, Ian and Dhebar, Yashesh and Deb, Kalyanmoy and Goodman, Erik D and Banzhaf, Wolfgang and Boddeti, Vishnu Naresh},
  journal={IEEE Transactions on Evolutionary Computation},
  volume={25},
  number={2},
  pages={277--291},
  year={2020},
  publisher={IEEE}
}

@inproceedings{CARS,
  title={Cars: Continuous evolution for efficient neural architecture search},
  author={Yang, Zhaohui and Wang, Yunhe and Chen, Xinghao and Shi, Boxin and Xu, Chao and Xu, Chunjing and Tian, Qi and Xu, Chang},
  booktitle={Proceedings of the IEEE/CVF conference on computer vision and pattern recognition},
  pages={1829--1838},
  year={2020}
}

@article{NPENAS,
  title={Npenas: Neural predictor guided evolution for neural architecture search},
  author={Wei, Chen and Niu, Chuang and Tang, Yiping and Wang, Yue and Hu, Haihong and Liang, Jimin},
  journal={IEEE Transactions on Neural Networks and Learning Systems},
  volume={34},
  number={11},
  pages={8441--8455},
  year={2022},
  publisher={IEEE}
}

@article{huang2022particle,
  title={Particle swarm optimization for compact neural architecture search for image classification},
  author={Huang, Junhao and Xue, Bing and Sun, Yanan and Zhang, Mengjie and Yen, Gary G},
  journal={IEEE Transactions on Evolutionary Computation},
  volume={27},
  number={5},
  pages={1298--1312},
  year={2022},
  publisher={IEEE}
}

@article{yuan2023particle,
  title={Particle swarm optimization for efficiently evolving deep convolutional neural networks using an autoencoder-based encoding strategy},
  author={Yuan, Gonglin and Wang, Bin and Xue, Bing and Zhang, Mengjie},
  journal={IEEE Transactions on Evolutionary Computation},
  year={2023},
  publisher={IEEE}
}

@article{han2024sadenas,
  title={SaDENAS: A self-adaptive differential evolution algorithm for neural architecture search},
  author={Han, Xiaolong and Xue, Yu and Wang, Zehong and Zhang, Yong and Muravev, Anton and Gabbouj, Moncef},
  journal={Swarm and Evolutionary Computation},
  volume={91},
  pages={101736},
  year={2024},
  publisher={Elsevier}
}

@article{xue2023neural,
  title={Neural architecture search based on a multi-objective evolutionary algorithm with probability stack},
  author={Xue, Yu and Chen, Chen and S{\l}owik, Adam},
  journal={IEEE Transactions on Evolutionary Computation},
  volume={27},
  number={4},
  pages={778--786},
  year={2023},
  publisher={IEEE}
}

@article{yang2023sweet,
  title={Sweet Gradient matters: Designing consistent and efficient estimator for Zero-shot Architecture Search},
  author={Yang, Longxing and Fu, Yanxin and Lu, Shun and Sun, Zihao and Mei, Jilin and Zhao, Wenxiao and Hu, Yu},
  journal={Neural Networks},
  volume={168},
  pages={237--255},
  year={2023},
  publisher={Elsevier}
}

@article{cai2018proxylessnas,
  title={Proxylessnas: Direct neural architecture search on target task and hardware},
  author={Cai, Han and Zhu, Ligeng and Han, Song},
  journal={arXiv preprint arXiv:1812.00332},
  year={2018}
}

@article{shu2021nasi,
  title={NASI: Label-and data-agnostic neural architecture search at initialization},
  author={Shu, Yao and Cai, Shaofeng and Dai, Zhongxiang and Ooi, Beng Chin and Low, Bryan Kian Hsiang},
  journal={arXiv preprint arXiv:2109.00817},
  year={2021}
}

@article{sun2022entropy,
  title={Entropy-driven mixed-precision quantization for deep network design},
  author={Sun, Zhenhong and Ge, Ce and Wang, Junyan and Lin, Ming and Chen, Hesen and Li, Hao and Sun, Xiuyu},
  journal={Advances in Neural Information Processing Systems},
  volume={35},
  pages={21508--21520},
  year={2022}
}

@inproceedings{liu2024evolution,
  title={Evolution of heuristics: Towards efficient automatic algorithm design using large language model},
  author={Liu, Fei and Tong, Xialiang and Yuan, Mingxuan and Lin, Xi and Luo, Fu and Wang, Zhenkun and Lu, Zhichao and Zhang, Qingfu},
  booktitle={Proceedings of the 41st International Conference on Machine Learning},
  series={Proceedings of Machine Learning Research},
  volume={235},
  pages={32201--32223},
  year={2024},
  url={https://proceedings.mlr.press/v235/liu24bs.html}
}

@article{glm2024chatglm,
  title={Chatglm: A family of large language models from glm-130b to glm-4 all tools},
  author={Glm, Team and Zeng, Aohan and Xu, Bin and Wang, Bowen and Zhang, Chenhui and Yin, Da and Zhang, Dan and Rojas, Diego and Feng, Guanyu and Zhao, Hanlin and others},
  journal={arXiv preprint arXiv:2406.12793},
  year={2024}
}

@article{deepseek2026deepseek,
  title={DeepSeek-V4: Towards Highly Efficient Million-Token Context Intelligence},
  author={DeepSeek-AI},
  journal={https://huggingface. co/deepseek-ai/DeepSeek-V4-Pro/blob/main/DeepSeek\_V4. pdf},
  year={2026}
}

@article{huang2025evorep,
  title={EvoREP: Evolving Reliable Ensemble of Proxies for Zero-Shot Neural Architecture Search},
  author={Huang, Junhao and Xue, Bing and Sun, Yanan and Zhang, Mengjie and Yen, Gary G.},
  journal={IEEE Transactions on Evolutionary Computation},
  year={2025},
  doi={10.1109/TEVC.2025.3646736}
}

@article{lai2026llmenas,
  title={LLMENAS: Evolutionary Neural Architecture Search via Large Language Model Guidance},
  author={Lai, Yutao and Cai, Zicheng and Chen, Lei and Ling, Tongtao and Liu, Hai-Lin},
  journal={IEEE Transactions on Evolutionary Computation},
  year={2026},
  publisher={IEEE}
}

@article{ma2026llamoco,
title={Llamoco: Instruction tuning of large language models for optimization code generation},
author={Ma, Zeyuan and Gong, Yue-Jiao and Guo, Hongshu and Chen, Jiacheng and Ma, Yining and Cao, Zhiguang and Zhang, Jun},
journal={IEEE Transactions on Evolutionary Computation},
year={2026},
publisher={IEEE}
}

@article{wang2026evolution,
title={Evolution of Benchmark: Black-Box Optimization Benchmark Design through Large Language Model},
author={Wang, Chen and Ma, Sijie and Ma, Zeyuan and Gong, Yue-Jiao},
journal={arXiv preprint arXiv:2601.21877},
year={2026}
}

@article{huang2025evaluation,
title={Evaluation of large language models as solution generators in complex optimization},
author={Huang, Beichen and Wu, Xingyu and Zhou, Yu and Wu, Jibin and Feng, Liang and Cheng, Ran and Tan, Kay Chen},
journal={IEEE Computational Intelligence Magazine},
volume={20},
number={4},
pages={56--70},
year={2025},
publisher={IEEE}
}

\end{document}